\documentclass{article}

\usepackage{iclr2026_conference_custom}

\usepackage{amsmath,amsfonts,bm}

\def\eqref#1{equation~\ref{#1}}

\def\1{\bm{1}}

\DeclareMathAlphabet{\mathsfit}{\encodingdefault}{\sfdefault}{m}{sl}
\SetMathAlphabet{\mathsfit}{bold}{\encodingdefault}{\sfdefault}{bx}{n}

\usepackage{graphicx}
\usepackage{booktabs}
\usepackage{subcaption}
\usepackage{multirow}
\usepackage{amsmath}
\usepackage[table]{xcolor}

\definecolor{steelblue}{RGB}{50,100,145}

\usepackage{hyperref}

\hypersetup{
    colorlinks=true,
    linkcolor=steelblue,          
    citecolor=steelblue,    
    urlcolor=steelblue      
}

\newtheorem{theorem}{Theorem}

\title{Geometry of Forgetting: Representation Flux in Continual Learning}
\author{
Maksim A. Kazanskii$^{*}$\\
$^{*}$\texttt{ mkazanskii@gmail.com}
}
\date{July 2026}

\begin{document}

\maketitle

\begin{abstract}

Catastrophic forgetting remains a fundamental obstacle to continual
learning, where neural networks lose previously acquired knowledge while
learning new tasks. Existing methods primarily mitigate forgetting through
parameter regularization or experience replay, yet the representation-space
dynamics associated with forgetting remain less well understood. In this
work, we investigate the evolution of latent representations during
sequential learning and introduce \emph{representation flux}, a geometric
quantity that measures sample-level representation displacement across
training. We show that representation flux is strongly associated with
catastrophic forgetting across multiple continual learning benchmarks,
while temporal analyses provide evidence that elevated flux can precede
subsequent performance degradation. Larger representation displacement is
also associated with greater confidence degradation, and complementary
geometric properties of representation transitions provide additional
information about sample-level forgetting. Motivated by these observations,
we propose \emph{FlowLess-R}, a simple representation-space regularization
method that constrains replay-sample representations relative to stored
reference representations while allowing continued learning of new tasks. FlowLess-R is architecture-agnostic and can be integrated into existing
replay-based continual learning methods by adding a representation-matching
term to the training objective. Experiments on SplitMNIST, SplitFashionMNIST, SplitCIFAR10, and
SplitTinyImageNet demonstrate improvements in final average accuracy and reductions in
catastrophic forgetting when FlowLess-R is combined with ER, DER++, and
ER-ACE. Together, our results identify representation flux as an informative
geometric marker of forgetting and show that stabilizing latent
representations provides a simple and effective strategy for mitigating
catastrophic forgetting.

\end{abstract}
\section{Introduction}

Catastrophic forgetting remains a fundamental obstacle to continual
learning \citep{wang2024survey,delange2021continual}. When neural networks are trained sequentially on multiple
tasks, performance on previously learned tasks often deteriorates despite
successful optimization of the current task. Existing continual learning methods primarily mitigate forgetting through parameter
regularization \citep{kirkpatrick2017ewc,zenke2017si,aljundi2018mas},
experience replay \citep{rebuffi2017icarl,rolnick2019experience,buzzega2020dark,chaudhry2019er},
optimization strategies \citep{lopez2017gradient},
or architectural modifications \citep{rusu2016progressive,serra2018hat,mallya2018packnet}.
While effective, these approaches provide limited insight into the geometric
processes occurring within the learned representation space that underlie
catastrophic forgetting.

In this work, we adopt a representation-space perspective on continual
learning. We view the latent representations as an evolving geometric distribution whose structure changes throughout sequential training. Rather than focusing solely on parameter updates, we study how
representations themselves move over time. This viewpoint naturally leads
to the notion of \emph{representation flux}, which measures the magnitude
of latent-space displacement between consecutive training epochs.

Our empirical analysis reveals a strong relationship between representation
flux and catastrophic forgetting across multiple continual learning
benchmarks. Temporal analyses provide evidence that elevated representation
flux can precede subsequent performance degradation, while sample-level
analyses show that larger representation displacements are associated with
higher forgetting rates and confidence degradation. Motivated by these
observations, we introduce \textbf{FlowLess-R} (\textbf{FlowLess} for
\textbf{R}epresentations), which constrains representation
displacement in latent space.

We provide a unified sample-level
geometric characterization of representation dynamics and catastrophic
forgetting, combining temporal representation displacement with local
density and transition-level geometric descriptors. We show that these
quantities characterize complementary aspects of sample-level forgetting
and relate representation displacement to changes in prediction confidence.
Motivated by these observations, we introduce FlowLess-R, a simple representation-space regularization
method that constrains replay-sample representations relative to stored
reference representations while allowing continued learning of new tasks. Finally, we demonstrate that
FlowLess-R improves multiple replay-based continual learning methods across
datasets, architectures, and replay-buffer capacities.

\section{Related Work}

\paragraph{Continual Learning.}

Continual learning methods are commonly categorized into
regularization-based, replay-based, optimization-based, and
architectural approaches.
Regularization methods preserve parameters that are important for
previous tasks using importance estimates such as the Fisher
information or path-integral measures
\citep{kirkpatrick2017ewc,zenke2017si,aljundi2018mas}.
Replay methods alleviate forgetting by rehearsing stored or generated
examples and currently represent one of the strongest and most widely
used continual learning paradigms
\citep{rebuffi2017icarl,rolnick2019experience,buzzega2020dark,chaudhry2019er}.
Optimization-based methods explicitly reduce gradient interference
between tasks
\citep{lopez2017gradient},
whereas architectural methods allocate task-specific network resources
to avoid destructive interference
\citep{rusu2016progressive,serra2018hat,mallya2018packnet}.
Unlike these approaches, our work focuses on the geometric dynamics of
latent representations and proposes a representation-space
regularizer that can be incorporated into existing replay methods.

\paragraph{Representation dynamics and stabilization.}
A growing body of continual learning research has examined changes in learned
representations as a source of catastrophic forgetting. Less-Forgetful
Learning directly constrains penultimate-layer features of the current model
to remain close to those produced by the previous model
\citep{jung2018less}, demonstrating that feature preservation can mitigate
forgetting. Subsequent approaches have developed more structured forms of
representation distillation. PODNet preserves intermediate representations
through spatially pooled feature distillation
\citep{douillard2020podnet}, while Adaptive Feature Consolidation estimates
the importance of individual feature maps and selectively restricts changes
to features that are important for preserving previously acquired knowledge
\citep{kang2022adaptive}. Related work has also investigated representation
change itself during continual learning. In particular, Caccia et
al.~\citep{caccia2022new} showed that abrupt representation changes at task
boundaries can produce substantial interference in online continual learning
and proposed ER-ACE, which mitigates this effect through an asymmetric
classification update that encourages new classes to adapt to previously
learned representations rather than the reverse. FER \citep{yan2022learning} similarly stores historical representations of
replay samples and uses them as additional supervision. FlowLess-R differs
primarily in its motivation and formulation: it derives representation
anchoring from an analysis of representation flux and its relationship to
catastrophic forgetting.

Our work differs in its primary focus. Rather than introducing representation
matching as a regularization mechanism, we study representation displacement
as a temporal, sample-level phenomenon during continual learning. We define
representation flux as the displacement of individual samples between
successive training snapshots and examine its relationship with subsequent
forgetting, confidence degradation, and other geometric properties of the
representation space. 

Motivated by these observations, we introduce FlowLess-R, a simple
stored-reference regularizer that penalizes the displacement of replay samples
from fixed, sample-specific historical representations stored when the samples
enter memory. Unlike previous-model feature distillation, this construction
uses persistent per-sample references and therefore directly constrains
accumulated representation drift without requiring a frozen teacher network.
FlowLess-R thus provides a direct intervention for testing whether limiting the
representation displacement identified in our analysis improves knowledge
retention.

\paragraph{Optimal Transport and Dynamical Systems.}

Optimal transport provides a principled framework for describing
probability distributions and their evolution
\citep{villani2009optimal,peyre2019computational}, and related
continuous-transport ideas have recently influenced generative modeling
\citep{lipman2023flow}.
The mathematical theory of gradient flows provides a related framework
for evolving probability measures \citep{ambrosio2008gradient}, while
dynamical-systems interpretations have also been developed for neural
networks \citep{chen2018neural}.
Our work differs from these approaches by studying continual learning
through the evolution of latent representation densities. Rather than optimizing transport itself, we characterize representation flux
as an informative geometric marker of forgetting and use this observation to
motivate regularization of representation dynamics during sequential learning.

\section{Representation Flux as an Early Indicator of Forgetting}

Before introducing our continual learning algorithm, we investigate the
geometric dynamics of latent representations during sequential learning.
Our goal is to determine whether changes in representation geometry provide
early indicators of catastrophic forgetting. Specifically, we ask the following question:

\begin{quote}
\emph{How is latent representation displacement related to subsequent
catastrophic forgetting during continual learning?}
\end{quote}

To answer this question, we analyze the trajectory of every training sample
throughout continual learning and relate its representation dynamics to
subsequent changes in prediction correctness, confidence, and forgetting.

\subsection{Definition of Representation Flux}

Let

\[
z_t=f_{\theta_t}(x)
\]

denote the latent representation of sample $x$ after training epoch $t$.
We define the \emph{representation flux} of a sample as the magnitude of
its representation displacement between consecutive training epochs,

\[
\Phi(x,t)=\|z_{t+1}-z_t\|_2,
\]

where larger values correspond to greater displacement in latent space. Intuitively, representation flux quantifies how strongly learning during
one training epoch alters previously learned representations. Small
representation flux indicates that the latent representation remains
geometrically stable across consecutive epochs, whereas large
representation flux reflects substantial reorganization of the latent
space.

\subsection{Representation Flux and Forgetting}

We first investigate the relationship between representation flux and
catastrophic forgetting. Unless otherwise stated, all analyses in this
section use SplitMNIST \citep{zenke2017si} with Experience Replay
(ER) \citep{chaudhry2019er}, five tasks, and a replay buffer of 40 samples
per task. During training, latent
representations are extracted after every training epoch from the last
hidden layer for all replay and validation samples. Representation flux, local representation density, prediction confidence,
and forgetting statistics are then computed independently for every sample;
the complete definitions are provided in
Appendix~\ref{app:additional_analysis}.

Figure~\ref{fig:flux_observation}(a) shows the temporal evolution of mean
representation flux together with the hard forgetting rate, based on the
forgetting-event definition of \citet{toneva2018example}, and the soft
forgetting score, measured as the decrease in predicted probability assigned
to the ground-truth class between consecutive epochs. Representation flux
peaks sharply at each task transition, whereas hard forgetting reaches its
local maximum one epoch later. Soft forgetting shows a qualitatively similar
but less consistent temporal pattern.

To quantify this relationship at the sample level,
Figure~\ref{fig:flux_observation}(b) aggregates statistics from all tasks
and training epochs of a representative continual learning run. Samples are grouped according to their
representation density and representation flux, and each bin reports the
empirical probability that a previously correct prediction becomes
incorrect after subsequent training. A clear trend emerges: the probability
of forgetting remains low for samples with small representation flux but
increases substantially as representation flux grows. Although local representation density is also associated with forgetting,
its relationship is considerably weaker than that observed for
representation flux. Even samples located
in dense regions of the latent space exhibit a high probability of
forgetting when their representations undergo sufficiently large
displacements.

Together, these observations identify representation flux as an informative
geometric marker of catastrophic forgetting and motivate the
representation-space regularization introduced in the next section.
Analogous trends are observed across SplitFashionMNIST
\citep{xiao2017fashion}, SplitCIFAR10 \citep{krizhevsky2009learning},
and SplitTinyImageNet \citep{le2015tiny}
(Appendix~\ref{app:additional_analysis}), although the temporal separation
between representation flux and forgetting is less visually pronounced
due to the coarser epoch-level discretization of training.

\begin{figure*}[t]
    \centering

    \begin{subfigure}[t]{0.48\textwidth}
        \centering
        \includegraphics[height=5.0cm]{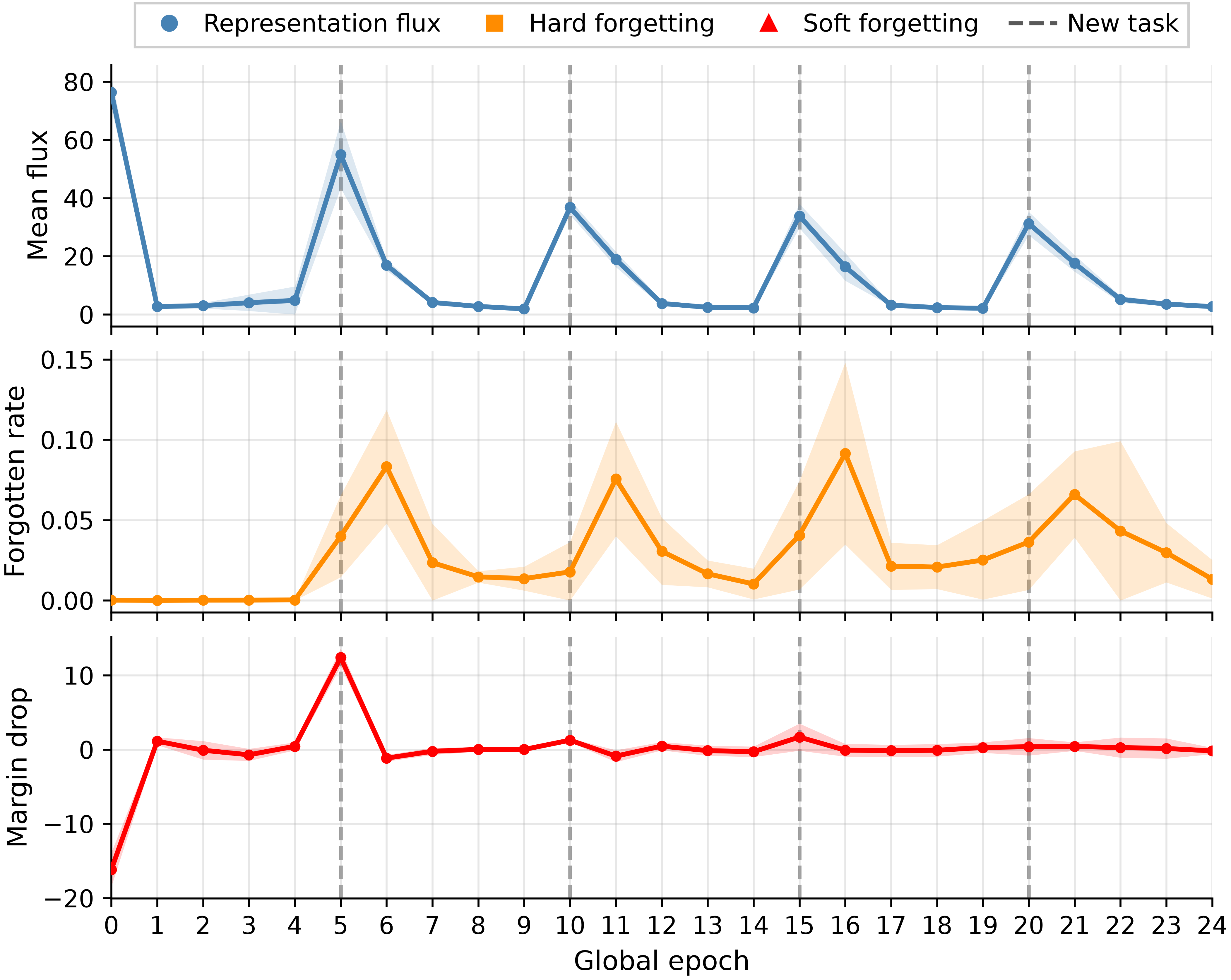}
        \caption{Representation flux tends to precede hard forgetting at task
        transitions, with a weaker temporal relationship for soft forgetting.}
        \label{fig:flux_dynamics}
    \end{subfigure}
    \hfill
    \begin{subfigure}[t]{0.48\textwidth}
        \centering
        \includegraphics[height=5.0cm]{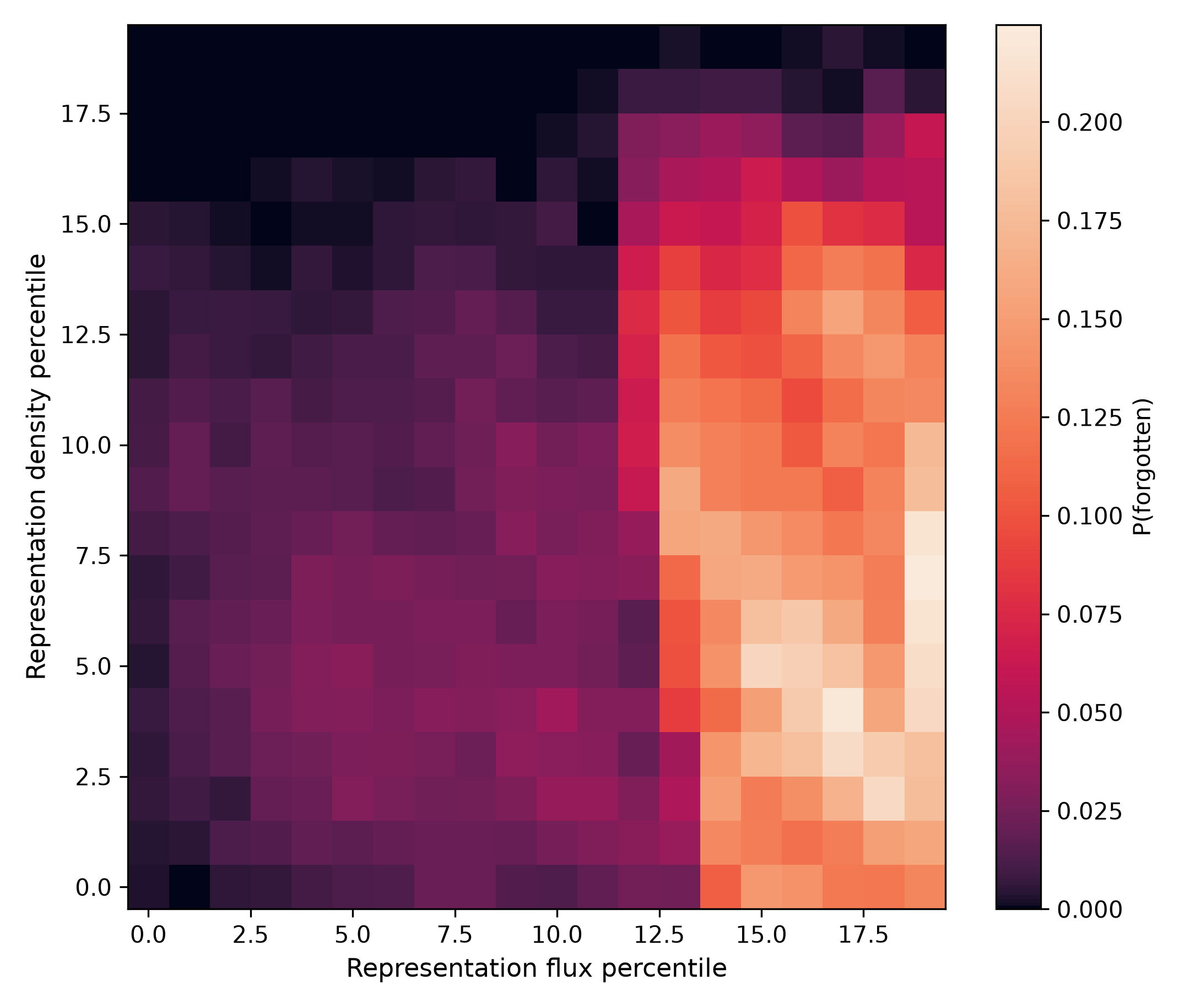}
        \caption{Probability of forgetting as a function of representation density and representation flux.}
        \label{fig:flux_heatmap}
    \end{subfigure}

    \caption{
    \textbf{Representation flux is associated with subsequent catastrophic forgetting.}
    \textbf{(a)} Temporal dynamics show that increases in representation
    flux tend to precede peaks in hard forgetting at task boundaries, while
    soft forgetting exhibits a qualitatively similar but weaker temporal
    pattern.
    \textbf{(b)} The probability of forgetting increases with
    representation flux, whereas representation density alone shows a
    weaker association. Together, these observations suggest that
    representation flux may serve as an early geometric indicator of
    subsequent catastrophic forgetting.
}
    \label{fig:flux_observation}
\end{figure*}

\subsection{Flux and Confidence Dynamics}

\begin{figure*}[t]
    \centering

    \begin{subfigure}[t]{0.32\textwidth}
        \centering
        \includegraphics[width=\linewidth]{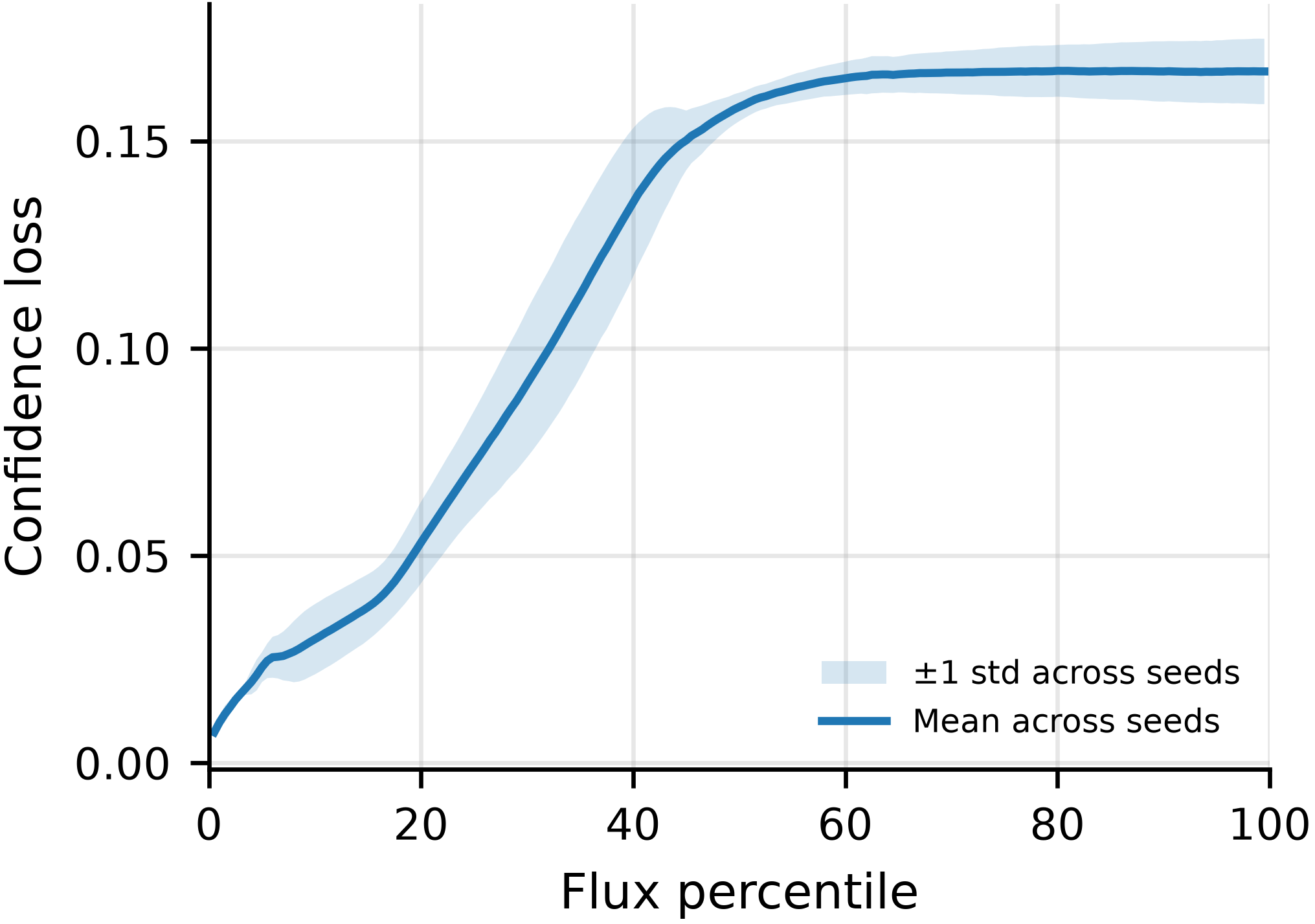}
        \caption{Average confidence loss versus representation flux.}
        \label{fig:flux_loss}
    \end{subfigure}
    \hfill
    \begin{subfigure}[t]{0.32\textwidth}
        \centering
        \includegraphics[width=\linewidth]{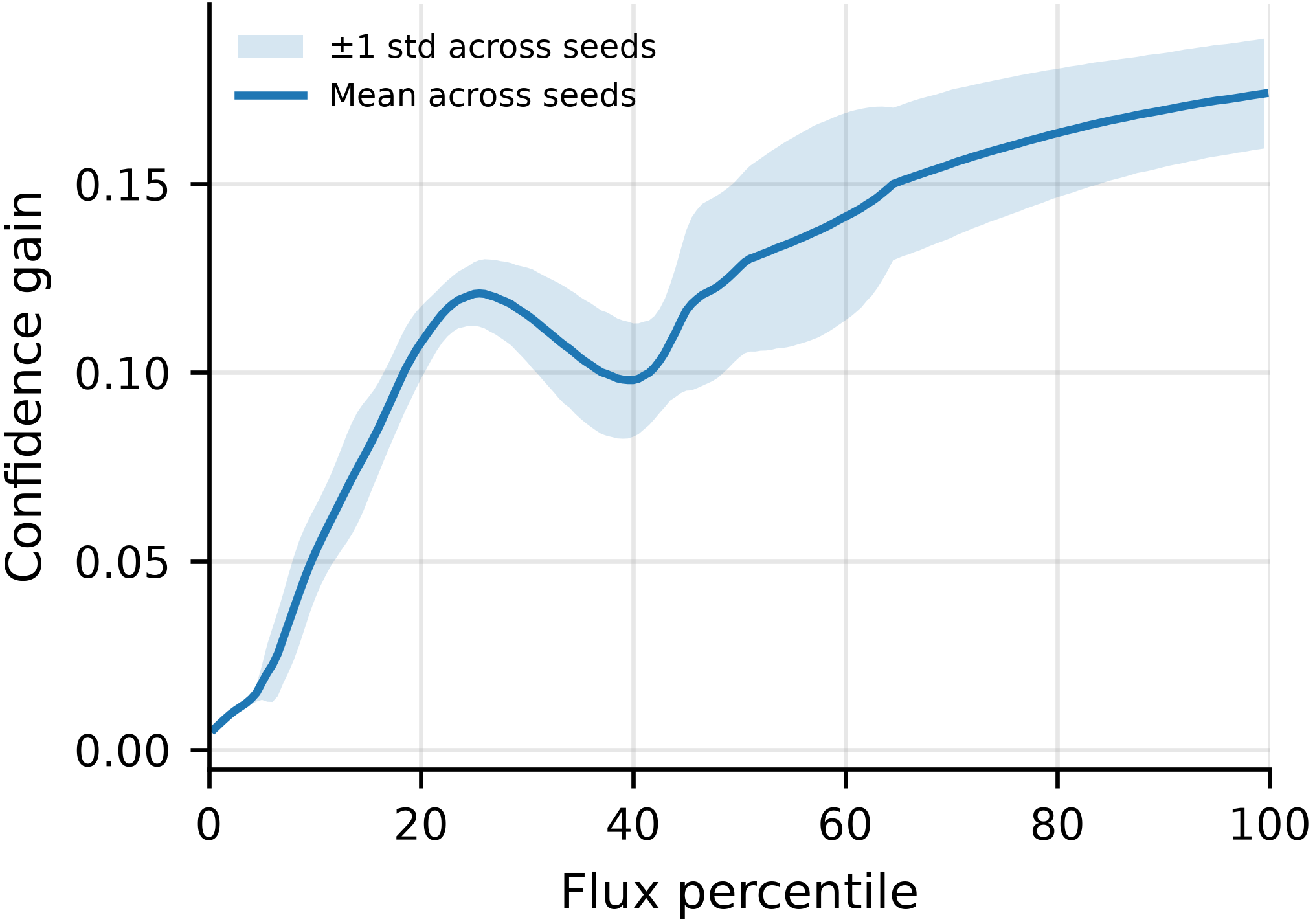}
        \caption{Average confidence gain versus representation flux.}
        \label{fig:flux_gain}
    \end{subfigure}
    \hfill
    \begin{subfigure}[t]{0.32\textwidth}
        \centering
        \includegraphics[width=\linewidth]{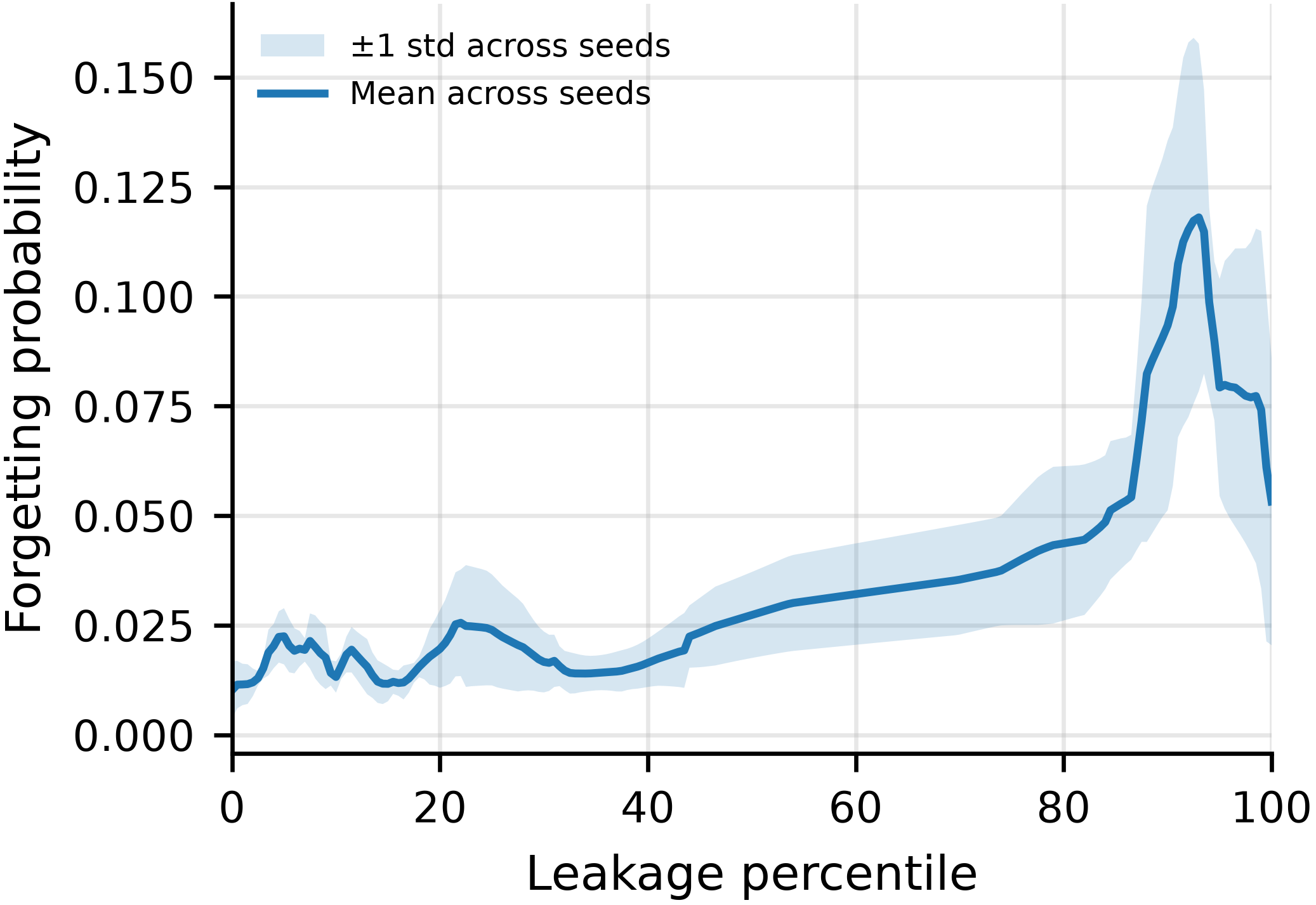}
        \caption{Average forgetting probability versus density leakage.}
        \label{fig:density_forgetting}
    \end{subfigure}

    \caption{
    \textbf{Representation flux characterizes confidence dynamics during
    continual learning.}
    \textbf{(a)} Confidence loss increases monotonically with representation
flux, with larger representation displacement associated with greater
confidence degradation.
    \textbf{(b)} Confidence gain is concentrated at substantially lower
levels of representation flux, whereas larger displacement is
predominantly associated with confidence degradation.
    \textbf{(c)} Forgetting probability generally increases with density
leakage, with a pronounced increase among samples exhibiting the
largest leakage values.
    }
    \label{fig:flux_confidence}
\end{figure*}

To better understand the relationship between representation flux and model
behavior, we analyze changes in prediction confidence together with a broader
set of geometric metrics. Figure~\ref{fig:flux_confidence}(a) shows that
confidence loss increases monotonically with representation flux: samples
undergoing larger representation displacements exhibit larger decreases in
prediction confidence. Conversely, Figure~\ref{fig:flux_confidence}(b) shows
that confidence gain is concentrated at substantially lower levels of flux.
These observations suggest that moderate representation displacement can
accompany confidence improvements, whereas larger displacement is
predominantly associated with confidence degradation.
Figure~\ref{fig:flux_confidence}(c) further shows that forgetting probability
generally increases with density leakage, which measures the tendency of
representations to leave their initial regions of a fixed latent-space
partition between consecutive representation snapshots (see
Appendix~\ref{app:theory_stability} for the formal definition).

We next examine whether geometric properties of the latent representation
are informative of sample-level forgetting. For each transition between consecutive training epochs, we characterize
representation changes using representation flux, local representation
density, density change, density leakage, transition stability, and
transition entropy; detailed definitions of these quantities are provided
in Appendix~\ref{app:additional_analysis}. We then train logistic
regression classifiers using different combinations of these geometric
features to distinguish samples that are forgotten during the corresponding
transition. Performance is measured using the area under the ROC curve (AUC),
computed independently for each random seed and summarized across runs.

Table~\ref{tab:forgetting_auc} summarizes the results across continual
learning benchmarks. Representation flux alone is substantially more
informative of sample-level forgetting than representation density on
SplitMNIST \citep{zenke2017si}, SplitFashionMNIST
\citep{xiao2017fashion}, and SplitCIFAR10 \citep{krizhevsky2009learning}.
Combining flux with density further improves discrimination on these
benchmarks, indicating that density provides complementary information.
On SplitTinyImageNet, neither flux nor density is informative in isolation,
but their combination improves discrimination, with further gains obtained
from additional geometric descriptors. Using all geometric features yields
the highest AUC on all four benchmarks, reaching $0.783$, $0.740$, $0.796$,
and $0.629$ on SplitMNIST, SplitFashionMNIST, SplitCIFAR10, and
SplitTinyImageNet \citep{le2015tiny}, respectively. These results show that
representation flux is a strong individual discriminator of sample-level
forgetting on three of the four benchmarks, while richer geometric
descriptions provide additional information.

\begin{table*}[t]
\centering
\scriptsize
\caption{Discriminative performance for sample-level forgetting across
continual learning benchmarks. Logistic regression models use different
combinations of geometric features computed over consecutive training
snapshots to distinguish samples that are forgotten during the corresponding
transition. Performance is reported as area under the ROC curve
(AUC; mean $\pm$ standard deviation) over five continual learning runs for
SplitMNIST, SplitFashionMNIST, and SplitCIFAR10, and three runs for
SplitTinyImageNet.}
\label{tab:forgetting_auc}
\resizebox{\textwidth}{!}{
\begin{tabular}{lcccc}
\toprule
Features &
SplitMNIST &
SplitFashionMNIST &
SplitCIFAR10 &
SplitTinyImageNet \\
\midrule

Density
& $0.557 \pm 0.043$
& $0.547 \pm 0.021$
& $0.502 \pm 0.010$
& $0.516 \pm 0.026$ \\

Representation Flux
& $0.733 \pm 0.004$
& $0.700 \pm 0.008$
& $0.766 \pm 0.004$
& $0.514 \pm 0.004$ \\

Flux + Density
& $0.761 \pm 0.012$
& $0.730 \pm 0.009$
& $0.781 \pm 0.005$
& $0.574 \pm 0.046$ \\

Flux + Density + Leakage
& $0.768 \pm 0.015$
& $0.738 \pm 0.010$
& $0.792 \pm 0.008$
& $0.606 \pm 0.021$ \\

Flux + Density + Stability
& $0.768 \pm 0.015$
& $0.738 \pm 0.010$
& $0.792 \pm 0.008$
& $0.606 \pm 0.021$ \\

Flux + Density + Entropy
& $0.759 \pm 0.012$
& $0.731 \pm 0.010$
& $0.787 \pm 0.008$
& $0.570 \pm 0.071$ \\

\midrule

\textbf{All geometric features}
& $\mathbf{0.783 \pm 0.014}$
& $\mathbf{0.740 \pm 0.011}$
& $\mathbf{0.796 \pm 0.007}$
& $\mathbf{0.629 \pm 0.006}$ \\

\bottomrule
\end{tabular}}
\end{table*}

These empirical observations also admit a natural geometric interpretation.
In Appendix~\ref{app:theory}, we relate representation flux to the evolution
of latent representation densities through a density transition process,
providing additional geometric motivation for FlowLess-R.

\section{FlowLess-R}

The empirical analysis in the preceding section reveals a strong
relationship between representation displacement and catastrophic forgetting,
while the temporal analysis provides evidence that elevated representation
flux can precede subsequent performance degradation. Unlike parameter-space regularization methods, FlowLess-R operates directly
in representation space. Rather than constraining network weights, it
penalizes deviations of replay-sample representations from fixed historical
references while retaining flexibility to learn new tasks.

\subsection{Representation Flux Regularization}

Consider a replay mini-batch
\[
\mathcal B_r = \{x_i\}_{i=1}^{N_r}.
\]
For each replay sample \(x_i\), we compute and store its latent representation
when the sample is added to the replay buffer,
\[
\tilde z_i = f_{\theta_i}(x_i),
\]
where \(\theta_i\) denotes the model parameters at the time of insertion.
The reference representation \(\tilde z_i\) remains fixed while the sample
is stored in the replay buffer.

Although representation flux measures displacement between consecutive
training snapshots, a sequence of such displacements can accumulate into
substantial drift from an earlier representation. FlowLess-R controls this
accumulated drift by anchoring each replay sample to its fixed historical
reference. During subsequent training, the current representation is
\[
z_i = f_\theta(x_i).
\]

We define the representation flux regularization loss as

\[
\mathcal L_{\mathrm{flux}}
=
\frac{1}{N_r}
\sum_{i=1}^{N_r}
\|
z_i-\tilde z_i
\|_2^2.
\]

A density-weighted extension is examined in
Section~\ref{sec:density_weighting}; its limited effect supports the simpler
unweighted formulation used throughout the main experiments. The complete optimization objective is
\[
\mathcal L
=
\mathcal L_{\mathrm{CL}}
+
\lambda\mathcal L_{\mathrm{flux}},
\]
where \(\mathcal L_{\mathrm{CL}}\) denotes the original continual learning
objective (ER, DER++, or ER-ACE), and \(\lambda\) controls the strength of
representation stabilization. This formulation requires no architectural modifications and can be combined
directly with existing replay-based continual learning methods.

\subsection{Computational complexity.}

Let $B_r$ denote the replay batch size, $d$ the representation dimension,
and $M$ the replay-buffer capacity. FlowLess-R adds
$\mathcal{O}(B_r d)$ computation per training step, compared with the
standard forward/backward cost $\mathcal{O}(B_r C_{\mathrm{net}})$, where
$C_{\mathrm{net}}$ denotes the per-sample network computation. Since
typically $d \ll C_{\mathrm{net}}$, this additional computational cost is small relative to the standard network forward/backward cost. FlowLess-R additionally requires
$\mathcal{O}(Md)$ memory for storing fixed reference representations.
Thus, compared with ER, ER-ACE, and DER++, FlowLess-R preserves linear
memory scaling with replay-buffer capacity $M$, while increasing the
constant storage cost per replay sample.
\section{Results}

\begin{table*}[t]
\centering
\caption{Mean forgetting (\%, mean $\pm$ standard deviation) for different
values of the FlowLess-R regularization coefficient $\lambda$. Lower values
indicate better knowledge retention. Results are averaged over 10 random
seeds for SplitMNIST, SplitFashionMNIST, and SplitCIFAR10, and 5 random
seeds for SplitTinyImageNet. The setting $\lambda=0$ corresponds to the
baseline method without FlowLess-R regularization. Reported $p$-values are
Holm--Bonferroni corrected for multiple comparisons across $\lambda$
values.}
\label{tab:forgetting}
\resizebox{\textwidth}{!}{
\begin{tabular}{llcccc}
\toprule
Algorithm & $\lambda$ &
SplitMNIST &
SplitFashionMNIST &
SplitCIFAR10 &
SplitTinyImageNet \\
\midrule
\textbf{Backbone} & &
\textbf{MLP} &
\textbf{MLP} &
\textbf{ResNet-18} &
\textbf{ResNet-18} \\
\midrule
\textbf{Buffer / task} & &
\textbf{40} &
\textbf{40} &
\textbf{400} &
\textbf{800} \\
\midrule

\multirow{6}{*}{ER}
& 0.0 & 32.69 $\pm$ 1.79 & 41.54 $\pm$ 1.23 & 27.75 $\pm$ 4.39 & 56.85 $\pm$ 1.26 \\
\cmidrule(lr){2-6}
& 0.1 & 28.74 $\pm$ 1.58 & 37.98 $\pm$ 2.14 & 19.97 $\pm$ 5.64 & \textbf{50.89 $\pm$ 2.42} \\
& 0.3 & 24.98 $\pm$ 2.27 & 35.39 $\pm$ 1.10 & \textbf{19.34 $\pm$ 3.92} & 54.56 $\pm$ 1.73 \\
& 1.0 & \textbf{24.01 $\pm$ 1.63} & \textbf{33.86 $\pm$ 1.45} & 20.39 $\pm$ 3.35 & 56.74 $\pm$ 2.97 \\
& 3.0 & 25.85 $\pm$ 1.22 & 35.66 $\pm$ 2.28 & 21.23 $\pm$ 3.04 & 52.55 $\pm$ 2.25 \\
\cmidrule(lr){2-6}
& \textbf{Best $\Delta$} &
\textbf{-8.68 ($p<0.0001$)} &
\textbf{-7.68 ($p<0.0001$)} &
\textbf{-8.41 ($p=0.0042$)} &
\textbf{-5.96 ($p=0.0361$)} \\
\midrule

\multirow{6}{*}{DER++}
& 0.0 & 26.11 $\pm$ 2.38 & 32.31 $\pm$ 2.51 & 25.78 $\pm$ 8.30 & 45.09 $\pm$ 1.32 \\
\cmidrule(lr){2-6}
& 0.1 & 25.57 $\pm$ 2.30 & 32.41 $\pm$ 2.45 & 14.70 $\pm$ 2.61 & 30.42 $\pm$ 3.34 \\
& 0.3 & 26.23 $\pm$ 1.64 & 31.76 $\pm$ 2.19 & 13.92 $\pm$ 3.01 & 29.02 $\pm$ 1.03 \\
& 1.0 & 25.39 $\pm$ 1.10 & \textbf{31.59 $\pm$ 2.53} & 13.60 $\pm$ 2.20 & 26.84 $\pm$ 1.66 \\
& 3.0 & \textbf{24.81 $\pm$ 1.89} & 31.68 $\pm$ 2.64 & \textbf{12.17 $\pm$ 2.13} & \textbf{26.03 $\pm$ 3.10} \\
\cmidrule(lr){2-6}
& \textbf{Best $\Delta$}
& \textbf{-1.30 ($p=0.1455$)}
& \textbf{-0.72 ($p=0.3349$)}
& \textbf{-13.61 ($p=0.0021$)}
& \textbf{-19.07 ($p=0.0001$)} \\
\midrule

\multirow{6}{*}{ER-ACE}
& 0.0 & 32.70 $\pm$ 2.28 & 41.35 $\pm$ 3.10 & 25.96 $\pm$ 5.82 & 56.73 $\pm$ 1.04 \\
\cmidrule(lr){2-6}
& 0.1 & 29.42 $\pm$ 1.56 & 39.39 $\pm$ 3.51 & \textbf{17.19 $\pm$ 1.83} & 52.42 $\pm$ 1.68 \\
& 0.3 & 27.30 $\pm$ 1.25 & 36.61 $\pm$ 2.51 & 19.17 $\pm$ 2.60 & 54.23 $\pm$ 2.53 \\
& 1.0 & \textbf{26.54 $\pm$ 1.36} & \textbf{35.04 $\pm$ 1.77} & 19.11 $\pm$ 3.09 & 57.64 $\pm$ 0.87 \\
& 3.0 & 26.88 $\pm$ 1.75 & 35.10 $\pm$ 2.27 & 22.35 $\pm$ 4.85 & \textbf{51.88 $\pm$ 1.21} \\
\cmidrule(lr){2-6}
& \textbf{Best $\Delta$}
& \textbf{-6.17 ($p<0.0001$)}
& \textbf{-6.32 ($p<0.0001$)}
& \textbf{-8.78 ($p=0.0027$)}
& \textbf{-4.85 ($p=0.0017$)} \\

\bottomrule
\end{tabular}}
\end{table*}
\subsection{FlowLess-R Performance}

Table~\ref{tab:forgetting} reports mean forgetting for FlowLess-R across the
evaluated replay methods and benchmarks. FlowLess-R generally improves
knowledge retention, with statistically significant reductions in forgetting
across most evaluated settings. For ER, mean forgetting decreases by $8.68$
percentage points on SplitMNIST ($p<0.0001$), $7.68$ points on
SplitFashionMNIST ($p<0.0001$), $8.41$ points on SplitCIFAR10
($p=0.0042$), and $5.96$ points on SplitTinyImageNet ($p=0.0361$).
Thus, the reduction in forgetting with ER is statistically significant on
all four evaluated benchmarks.

For DER++, the largest reduction occurs on SplitTinyImageNet, where mean
forgetting decreases from $45.09\%$ to $26.03\%$, a reduction of $19.07$
percentage points ($p=0.0001$). A substantial reduction is also observed on
SplitCIFAR10, from $25.78\%$ to $12.17\%$, corresponding to $13.61$ points
($p=0.0021$). The smaller reductions on SplitMNIST and SplitFashionMNIST do
not reach statistical significance.

FlowLess-R also consistently improves retention when combined with ER-ACE.
Mean forgetting is reduced by $6.17$ points on SplitMNIST ($p<0.0001$),
$6.32$ points on SplitFashionMNIST ($p<0.0001$), $8.78$ points on
SplitCIFAR10 ($p=0.0027$), and $4.85$ points on SplitTinyImageNet
($p=0.0017$). Notably, these reductions remain statistically significant
across all four benchmarks and with both MLP and ResNet-18 backbones.

The reductions in forgetting are generally accompanied by improvements in
final average accuracy rather than a loss of predictive performance. In most
evaluated settings, FlowLess-R simultaneously reduces forgetting and improves
final average accuracy. Complete final average accuracy results, including
results across all evaluated values of $\lambda$, are reported in
Appendix~\ref{app:add_flow}.

\subsection{Density Weighting}

Our analysis indicates that representation flux and local representation
density provide complementary information about sample-level forgetting.
This observation motivates an optional density-weighted variant of
FlowLess-R, which weights replay samples according to their local
representation density during regularization. 

Appendix~\ref{sec:density_weighting} introduces the density-weighted formulation and evaluates its sensitivity to the weighting exponent $\alpha$. Although mildly negative values of $\alpha$ occasionally provide small improvements, the unweighted formulation ($\alpha=0$) achieves comparable performance across all replay buffer sizes while avoiding both an additional hyperparameter and the associated computational overhead. Consequently, the density-weighting term is omitted from all experiments in the main paper unless stated otherwise.

\subsection{Replay Buffer Size}

To evaluate the robustness of FlowLess-R across different replay capacities,
we varied the replay buffer size over dataset-specific ranges and compared
the resulting performance with the corresponding baseline replay methods.
The evaluated buffer sizes are specified in
Appendix~\ref{app:experimental_setup}, and
Appendix~\ref{sec:buffer_size} reports the complete results for replay
buffers ranging from very limited to substantially larger memory budgets.
FlowLess-R generally improves final average accuracy and reduces mean
forgetting across the evaluated replay capacities. On SplitMNIST and
SplitFashionMNIST, the gains are largest at smaller replay capacities and
gradually narrow as the buffer grows, whereas SplitCIFAR10 exhibits a less
pronounced improvement. In contrast, on SplitTinyImageNet the benefit
becomes particularly pronounced at larger replay capacities, where the
performance gap widens substantially and forgetting is markedly reduced.
One possible explanation is that larger buffers provide a broader set of
stored representation anchors in the more complex TinyImageNet
representation space, allowing FlowLess-R to constrain representation drift
more effectively. This interpretation remains speculative, however, as the
buffer-size experiments do not directly identify the mechanism underlying
the different scaling behavior across datasets. Overall, these results
suggest that FlowLess-R can complement experience replay across a broad
range of replay capacities rather than relying on a particular memory
budget.
\subsection{Layer Ablations}

To determine the most effective location for FlowLess-R regularization, we performed a layer ablation study on SplitMNIST using ER with FlowLess-R and a replay buffer of 40 samples per task. Table~\ref{tab:layers} reports results averaged over all regularization coefficients and random seeds (0--9). Regularizing the last hidden layer achieved the highest mean final average accuracy (78.30\%) and the lowest mean forgetting (25.90\%). Applying FlowLess-R to the last two hidden layers also improved performance relative to regularizing earlier layers. In this setting, these findings suggest that stabilizing later representations immediately before the classifier is more beneficial than constraining earlier representations or multiple layers simultaneously. Throughout this paper, FlowLess-R is applied to the last hidden layer unless stated otherwise. Complete implementation details, training protocols, hyperparameter settings, and evaluation procedures are provided in Appendix~\ref{app:experimental_setup}.

\begin{table}[t]
\centering
\scriptsize
\caption{Effect of applying the FlowLess-R regularizer to different network layers on SplitMNIST (ER, replay buffer of 40 samples per task). Results are averaged over all regularization coefficients and 10 random seeds (0--9). The last hidden layer provides the best overall performance and is therefore used throughout the paper.}
\label{tab:layers}
\begin{tabular}{lcc}
\toprule
Layers & Accuracy (\%) & Mean Forgetting (\%) \\
\midrule
Layer 1 & 72.77 $\pm$ 1.98 & 32.85 $\pm$ 2.51 \\
Layer 2 & 73.10 $\pm$ 1.49 & 32.49 $\pm$ 1.86 \\
\textbf{Layer 3} & \textbf{78.30 $\pm$ 1.93} & \textbf{25.90 $\pm$ 2.43} \\
Layer 2 + Layer 3 & 77.32 $\pm$ 2.62 & 27.07 $\pm$ 3.32 \\
Layer 1 + Layer 2 + Layer 3 & 75.76 $\pm$ 2.34 & 28.97 $\pm$ 3.05 \\
\bottomrule
\end{tabular}
\end{table}

\section{Discussion}

Our experiments provide evidence that catastrophic forgetting is closely linked to the dynamics of latent representations, suggesting that representation-space analysis provides information complementary to parameter-space descriptions. Large representation displacement is associated with
confidence degradation and higher forgetting rates, while our temporal
analysis provides evidence that elevated representation flux can precede
subsequent performance degradation. Together, these results suggest that
representation dynamics provide useful geometric information about
forgetting beyond prediction accuracy alone.

 The proposed FlowLess-R regularizer is motivated directly by this
observation. Instead of constraining network parameters, it stabilizes
replay-sample representations relative to their stored reference
representations. Despite its simplicity, this representation-space regularization improves knowledge retention across the evaluated replay methods and benchmarks, with particularly consistent reductions in forgetting, while requiring only an additional representation-matching term during training. The method is architecture-agnostic and requires only the additional storage of reference latent representations for replay samples, making it straightforward to integrate into existing replay-based continual learning algorithms.

Interestingly, the layer ablation study indicates that, in the evaluated setting, regularizing only the final hidden layer is more effective than constraining earlier layers or multiple layers simultaneously. A plausible explanation is that the last hidden layer more directly
supports the learned decision boundaries, whereas earlier feature-extraction
layers may benefit from greater flexibility during adaptation.

The improvements obtained when FlowLess-R is combined with ER-ACE are particularly informative because ER-ACE already mitigates abrupt representation change through its asymmetric classification objective. The additional gains from FlowLess-R suggest that directly constraining the accumulated displacement of replay-sample representations provides a complementary form of stability. ER-ACE reduces disruptive updates induced by incoming classes, whereas FlowLess-R anchors previously observed samples to persistent historical representations. Although these results do not establish a causal decomposition of the two mechanisms, they suggest that abrupt task-boundary interference and accumulated representation drift constitute complementary targets for continual learning regularization.

Several limitations remain. First, the present study focuses on
replay-based continual learning methods, and it remains to be investigated
whether representation-flux regularization is equally effective for
regularization-based or parameter-isolation approaches. Second, our
analysis considers image classification benchmarks; extending the proposed
framework to sequential decision making, language models, and multimodal
continual learning constitutes an important direction for future work.
Third, representation flux is also dependent on the geometry and scaling of the chosen latent space. Although several of the observed relationships remain qualitatively consistent across architectures and datasets, developing representation-invariant or normalized measures of latent displacement is an important direction for future work. Finally, while representation flux is shown to be an informative empirical
marker of forgetting, establishing theoretical guarantees relating
representation dynamics to generalization remains an open problem. The reduced informativeness of individual geometric descriptors on
SplitTinyImageNet, together with the improved discrimination obtained by
combining them, suggests that forgetting geometry may become increasingly
multivariate on more complex benchmarks. Investigating this behavior at
larger scales is an interesting direction for future work.

More broadly, our results suggest that continual learning can be viewed from a geometric perspective in which forgetting is associated with unstable evolution of latent representations. This viewpoint complements
parameter-space analyses and provides a simple principle for designing
future continual learning algorithms: rather than reacting to forgetting
after it occurs, explicitly control representation dynamics throughout
training.

\section{Conclusion}

We introduced a representation-space perspective on catastrophic
forgetting by studying the dynamics of latent representations during
continual learning. Our analyses identify representation flux as an
informative geometric marker of catastrophic forgetting. Motivated by
these observations, we proposed FlowLess-R, which directly stabilizes
latent representations of replay samples during training. Despite its
simplicity, this representation-space regularization improves knowledge
retention across the evaluated replay methods and benchmarks, with
particularly consistent reductions in forgetting, while requiring only
an additional representation-matching term during training. These results
suggest that controlling representation dynamics can provide an effective
and interpretable approach to mitigating catastrophic forgetting and
offers a promising direction for future continual learning research.

\clearpage
\bibliography{references} \bibliographystyle{iclr2026_conference} 

\clearpage
\tableofcontents
\clearpage
\appendix

\clearpage
\section{A Density Evolution Interpretation of Representation Flux}
\label{app:theory}

This appendix provides a theoretical interpretation of the empirical
observations presented in the main paper. Rather than viewing continual
learning solely through parameter updates, we consider the evolution of
latent representation densities during sequential learning. This viewpoint
naturally explains why excessive representation flux is associated with
catastrophic forgetting and motivates the proposed FlowLess-R
regularization.

\subsection{Representation Density Evolution}
\label{app:theory_density_evolution}
Let

\[
z=f_{\theta}(x)
\]

denote the latent representation of sample \(x\), where
\(f_{\theta}\) is the neural network parameterized by \(\theta\).
The distribution of latent representations after training epoch \(t\) is

\[
p_t(z).
\]

Between consecutive training snapshots, the network parameters evolve as

\[
\theta_t\rightarrow\theta_{t+1},
\]

thereby inducing a corresponding evolution of the latent representation
density,

\[
p_t(z)\rightarrow p_{t+1}(z).
\]

This viewpoint is inspired by geometric descriptions of evolving
probability measures in optimal transport, gradient flows, and
statistical physics
\citep{villani2009optimal,
peyre2019computational,
ambrosio2008gradient,
risken1996fokker}.

For an individual sample,

\[
z_t=f_{\theta_t}(x),
\qquad
z_{t+1}=f_{\theta_{t+1}}(x),
\]

while the collective motion of all samples induces the density transition
from \(p_t(z)\) to \(p_{t+1}(z)\).

This interpretation motivates FlowLess-R, which explicitly regularizes
representation dynamics by limiting unnecessary transport of previously
learned representations while preserving sufficient flexibility to learn
new tasks.

\subsection{Density Transition Matrix}
\label{app:theory_matrix}

To characterize the evolution of representation density
quantitatively, we introduce a finite discretization of the latent
space. Let

\[
r:\mathbb{R}^d\rightarrow\{1,\ldots,K\}
\]

be a region assignment function that maps each latent representation
to one of \(K\) regions. We consider a fixed measurable partition of
the latent space over each transition from \(t\) to \(t+1\), so that
the same region assignment is used for both representation snapshots.
The partition may be induced by clustering or a Voronoi tessellation
and serves solely as a mathematical discretization of the continuous
representation distribution.

Under this discretization, let

\[
p_t\in\mathbb{R}^{K}
\]

denote the vector whose \(i\)-th component is the probability mass
contained in region \(i\) at training epoch \(t\).

We define the \emph{Density Transition Matrix}

\[
M_t\in\mathbb{R}^{K\times K},
\]

whose entries are

\[
M_t(i,j)
=
\frac{
\left|
\left\{
n:
r(z_n^t)=i,\;
r(z_n^{t+1})=j
\right\}
\right|
}
{
\left|
\left\{
n:
r(z_n^t)=i
\right\}
\right|
}.
\]

Thus,

\[
M_t(i,j)
=
P\!\left(
r(z^{t+1})=j
\mid
r(z^t)=i
\right),
\]

is the empirical probability that a representation initially located
in region \(i\) moves to region \(j\) between consecutive training
snapshots. This construction is analogous to transition operators used in discrete-time Markov processes \citep{norris1998markov}.

Each row of \(M_t\) forms a probability distribution,

\[
M_t(i,j)\ge0,
\qquad
\sum_j M_t(i,j)=1.
\]

Diagonal entries quantify the fraction of representation density that
remains within the same latent-space region, whereas off-diagonal
entries measure transport of probability mass between different
regions. Consequently, the discretized density evolution is governed by

\[
p_{t+1}=M_t^{\top}p_t,
\]

which is analogous to the evolution equations of discrete transport
operators, Markov processes, and master equations in statistical
physics
\citep{norris1998markov,
vanKampen2007,
risken1996fokker,
villani2009optimal}.

The Density Transition Matrix therefore provides a compact
description of how optimization redistributes representation density
through the latent space. Large off-diagonal mass indicates
substantial transport of learned representations, whereas matrices
concentrated near the diagonal correspond to stable representation
geometry and limited latent-space drift.

\subsection{Density Stability}
\label{app:theory_stability}
The probability that a representation remains in its current latent region
is given by the diagonal entries of the Density Transition Matrix.
Accordingly, we define the density stability as

\[
S_t
=
\sum_{i=1}^{K}
p_t(i)\,M_t(i,i),
\]

where \(p_t(i)\) is the probability of region \(i\). The complementary
density leakage is

\[
\ell_t
=
1-S_t
=
\sum_{i=1}^{K}
p_t(i)\left(1-M_t(i,i)\right).
\]

Thus, density stability measures the probability that a randomly sampled
representation remains within its original latent region, whereas density
leakage measures the probability that it transitions to a different region.

\subsection{Geometric Bound Relating Representation Flux to Density Leakage}
\label{app:theory_bound}

Consider a fixed measurable partition of the latent space into regions
\(R_1,\ldots,R_K\), with region assignment function
\[
r:\mathbb{R}^d\rightarrow\{1,\ldots,K\}.
\]
For a sample \(x\), let
\[
i=r(z_t(x))
\]
denote its region at training snapshot \(t\), and define the
\emph{boundary clearance}
\[
B_t(x)
=
\operatorname{dist}\!\left(z_t(x),R_i^c\right),
\]
where \(R_i^c\) denotes the complement of \(R_i\). Thus, \(B_t(x)\)
is the minimum displacement required for \(z_t(x)\) to leave its
current region.

Let
\[
\Phi_t(x)
=
\|z_{t+1}(x)-z_t(x)\|_2
\]
denote the representation flux, and define the region-transition
indicator
\[
T_t(x)
=
\mathbf{1}
\left(
r(z_{t+1}(x))
\neq
r(z_t(x))
\right).
\]
The density leakage is then
\[
\ell_t
=
\Pr(T_t=1)
=
\sum_{i=1}^{K}
p_t(i)\left(1-M_t(i,i)\right).
\]

\begin{theorem}[Flux-Based Density Leakage Bound]
\label{thm:flux_leakage}

For any threshold \(\rho>0\),
\[
\ell_t
\le
\Pr(B_t\le\rho)
+
\Pr(\Phi_t\ge\rho).
\]
Consequently,
\[
\ell_t
\le
\inf_{\rho>0}
\left[
\Pr(B_t\le\rho)
+
\Pr(\Phi_t\ge\rho)
\right].
\]

\end{theorem}

\noindent\textbf{Proof.}

If a representation changes regions between snapshots \(t\) and \(t+1\),
its displacement must be at least as large as its boundary clearance.
Therefore,
\[
T_t
\le
\mathbf{1}(\Phi_t\ge B_t),
\]
and hence
\[
\ell_t
=
\mathbb{E}[T_t]
\le
\Pr(\Phi_t\ge B_t).
\]

For any \(\rho>0\), if \(\Phi_t\ge B_t\), then either
\(B_t\le\rho\) or \(\Phi_t\ge\rho\). Thus,
\[
\{\Phi_t\ge B_t\}
\subseteq
\{B_t\le\rho\}
\cup
\{\Phi_t\ge\rho\}.
\]
Applying the union bound gives
\[
\ell_t
\le
\Pr(B_t\le\rho)
+
\Pr(\Phi_t\ge\rho).
\]
Since this inequality holds for every \(\rho>0\), taking the infimum
over \(\rho\) yields the second result.

\hfill\(\square\)

The bound separates two sources of latent-space instability. The term
\(\Pr(B_t\le\rho)\) quantifies the mass of representations lying close
to the boundaries of their assigned regions and therefore depends on
the geometry of the latent representation distribution. The term
\(\Pr(\Phi_t\ge\rho)\) quantifies the frequency of sufficiently large
representation displacements and depends on the dynamics of the learned
representations.

This decomposition provides a geometric motivation for controlling
representation displacement. FlowLess-R penalizes deviations of replay
representations from their stored reference representations,
\[
\mathcal{L}_{\mathrm{flux}}
=
\frac{1}{N_r}
\sum_{n=1}^{N_r}
\|z_n-\tilde z_n\|_2^2,
\]
where \(\tilde z_n\) is the representation stored when sample \(x_n\)
is added to the replay buffer and \(z_n=f_\theta(x_n)\) is its current
representation. Although this fixed-reference displacement is distinct
from the consecutive-snapshot flux \(\Phi_t\) appearing in the bound,
FlowLess-R is designed to limit accumulated representation drift and
thereby stabilize the latent geometry of replay samples.

\subsection{Representation Flux and Decision Boundary Stability}
\label{app:theory_flux}
Let \(m_t(z)\) denote the classification margin at representation \(z\)
at training step \(t\), with positive margin indicating correct
classification. For a sample \(x\), let

\[
z_t(x)=f_{\theta_t}(x)
\]

denote its latent representation at step \(t\).

To distinguish changes in the classifier from changes in the
representation, define the classifier drift at the fixed representation
\(z_t(x)\) as

\[
\Delta_t(x)
=
\left|
m_{t+1}(z_t(x))
-
m_t(z_t(x))
\right|.
\]

Assume that the margin function \(m_{t+1}(\cdot)\) is
\(L_m\)-Lipschitz,

\[
|m_{t+1}(z_1)-m_{t+1}(z_2)|
\le
L_m\|z_1-z_2\|_2.
\]

Let

\[
\Phi_t(x)
=
\|z_{t+1}(x)-z_t(x)\|_2
\]

denote the representation flux, and define the forgetting event as

\[
F_t(x)
=
\mathbf{1}
\left(
m_t(z_t(x))>0,\;
m_{t+1}(z_{t+1}(x))\le0
\right).
\]

\begin{theorem}[Representation Flux Bounds Forgetting]
\label{thm:margin_flux}

If \(F_t(x)=1\), then

\[
L_m\Phi_t(x)+\Delta_t(x)
\ge
m_t(z_t(x)).
\]

Equivalently, whenever

\[
m_t(z_t(x))>\Delta_t(x),
\]

we have

\[
\Phi_t(x)
\ge
\frac{m_t(z_t(x))-\Delta_t(x)}{L_m}.
\]

\end{theorem}

\noindent\textbf{Proof.}

Decomposing the change in margin into classifier drift at the fixed
representation \(z_t(x)\) and the effect of representation displacement,
we obtain

\[
\begin{aligned}
m_{t+1}(z_{t+1}(x))
&\ge
m_{t+1}(z_t(x))
-
L_m\|z_{t+1}(x)-z_t(x)\|_2 \\
&\ge
m_t(z_t(x))
-
\Delta_t(x)
-
L_m\Phi_t(x).
\end{aligned}
\]

Hence, if

\[
L_m\Phi_t(x)+\Delta_t(x)
<
m_t(z_t(x)),
\]

then

\[
m_{t+1}(z_{t+1}(x))>0,
\]

contradicting \(F_t(x)=1\). Therefore,

\[
F_t(x)=1
\Longrightarrow
L_m\Phi_t(x)+\Delta_t(x)
\ge
m_t(z_t(x)).
\]

\hfill\(\square\)

Consequently,

\[
\Pr(F_t=1)
\le
\Pr\!\left(
L_m\Phi_t+\Delta_t
\ge
m_t(z_t)
\right).
\]

The theorem identifies two complementary mechanisms associated with
catastrophic forgetting: representation displacement, measured by
\(\Phi_t\), and classifier drift, measured by \(\Delta_t\).
Forgetting can occur only if their combined effect is sufficiently large
relative to the current classification margin. By constraining the
accumulated displacement of replay-sample representations from their
stored reference representations, FlowLess-R acts to suppress the
representation-drift component associated with \(\Phi_t\).
\clearpage

\section{Additional Representation Flux Analyses}
\label{app:additional_analysis}

To evaluate the generality of the proposed geometric framework, we repeat
the analyses from the main paper on SplitFashionMNIST
\citep{xiao2017fashion}, SplitCIFAR10 \citep{krizhevsky2009learning}, and
SplitTinyImageNet \citep{le2015tiny} using the same continual learning
protocol. The objective of these experiments is to determine whether the
relationships between representation dynamics, prediction confidence, and
catastrophic forgetting remain qualitatively consistent across datasets of
increasing visual complexity.

\begin{figure*}[t]
    \centering

    \textbf{SplitFashionMNIST}\\[2mm]

    \begin{subfigure}[t]{0.32\textwidth}
        \centering
        \includegraphics[width=\linewidth]{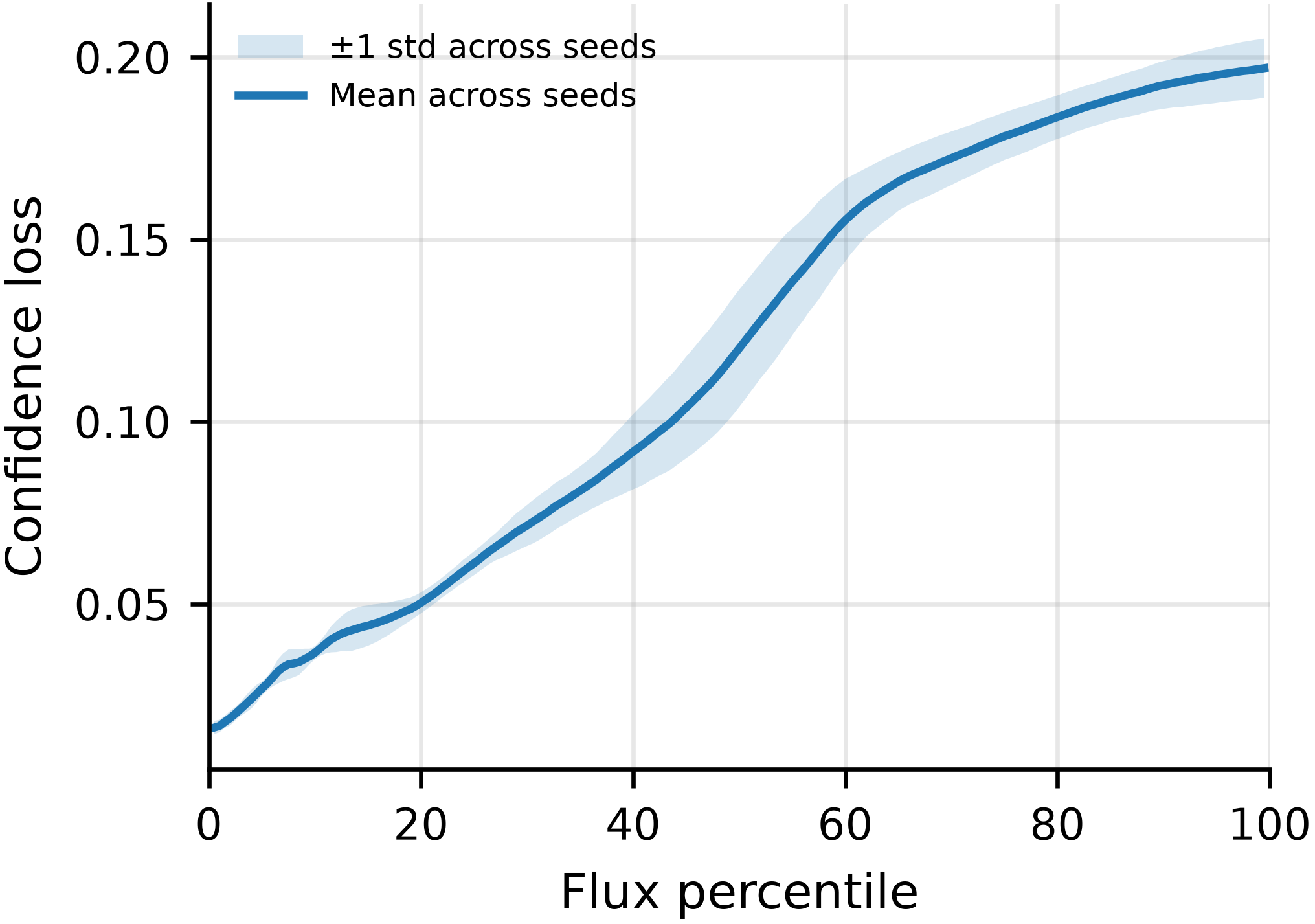}
    \end{subfigure}
    \hfill
    \begin{subfigure}[t]{0.32\textwidth}
        \centering
        \includegraphics[width=\linewidth]{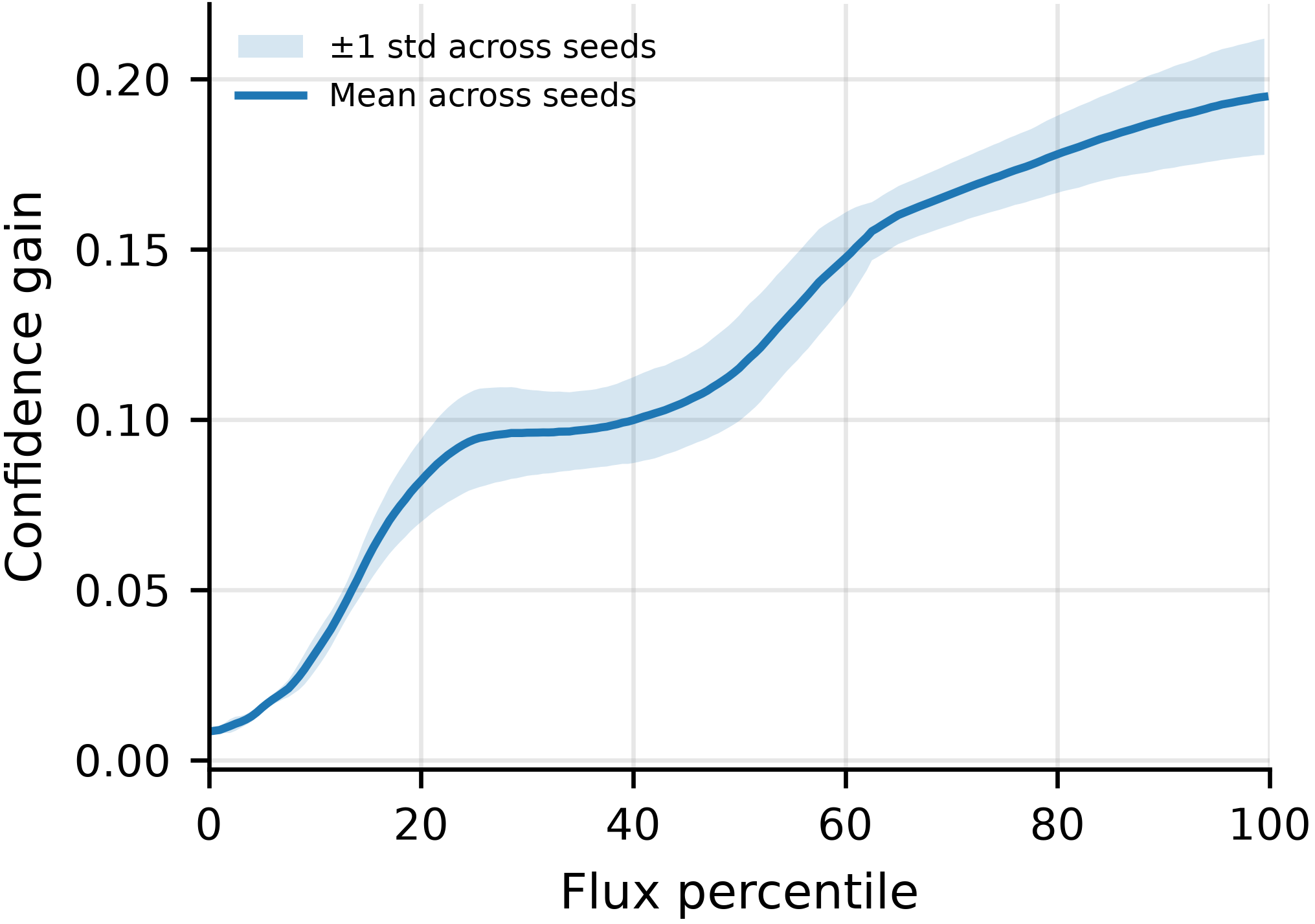}
    \end{subfigure}
    \hfill
    \begin{subfigure}[t]{0.32\textwidth}
        \centering
        \includegraphics[width=\linewidth]{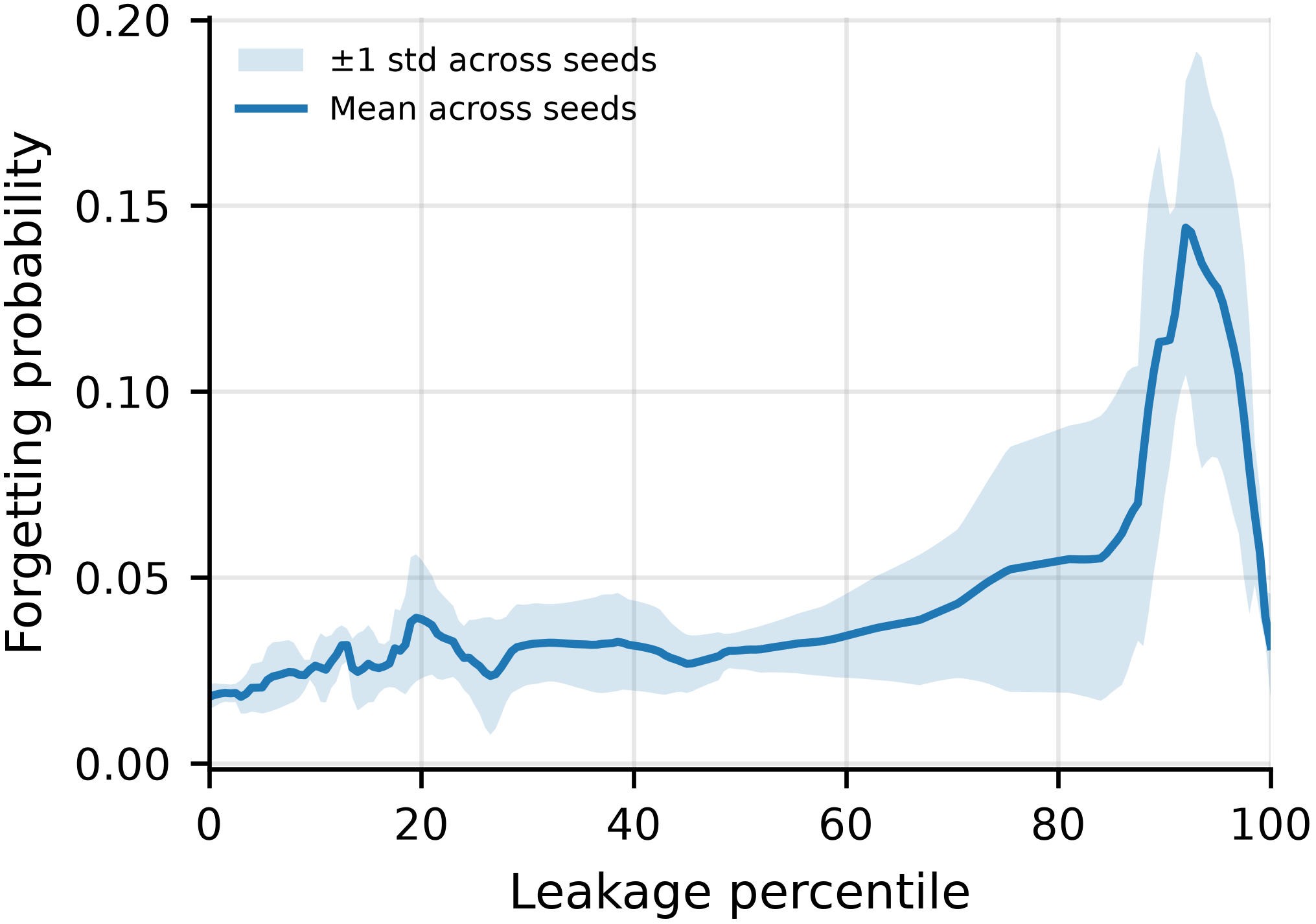}
    \end{subfigure}

    \vspace{3mm}

    \textbf{SplitCIFAR10}\\[2mm]

    \begin{subfigure}[t]{0.32\textwidth}
        \centering
        \includegraphics[width=\linewidth]{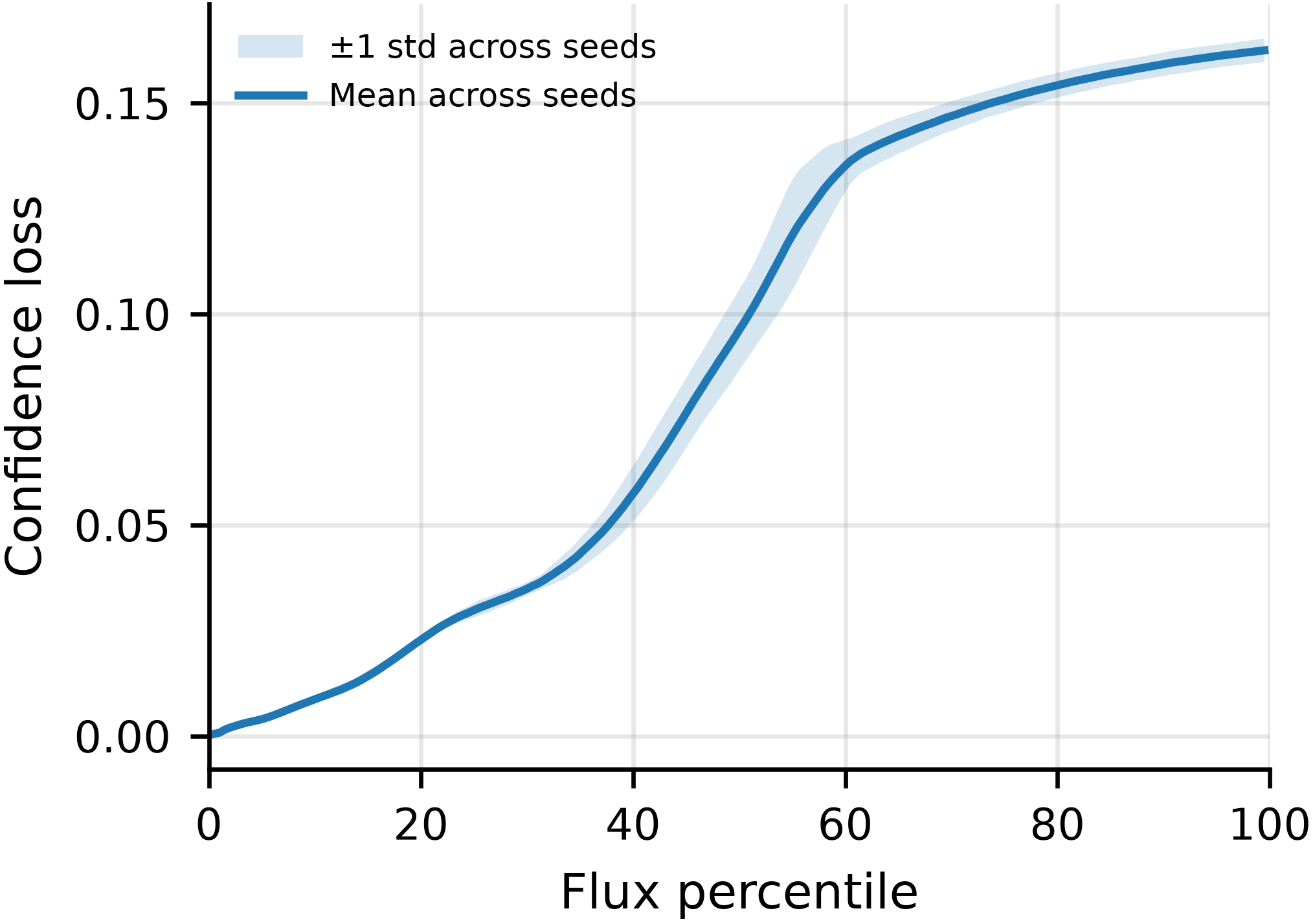}
    \end{subfigure}
    \hfill
    \begin{subfigure}[t]{0.32\textwidth}
        \centering
        \includegraphics[width=\linewidth]{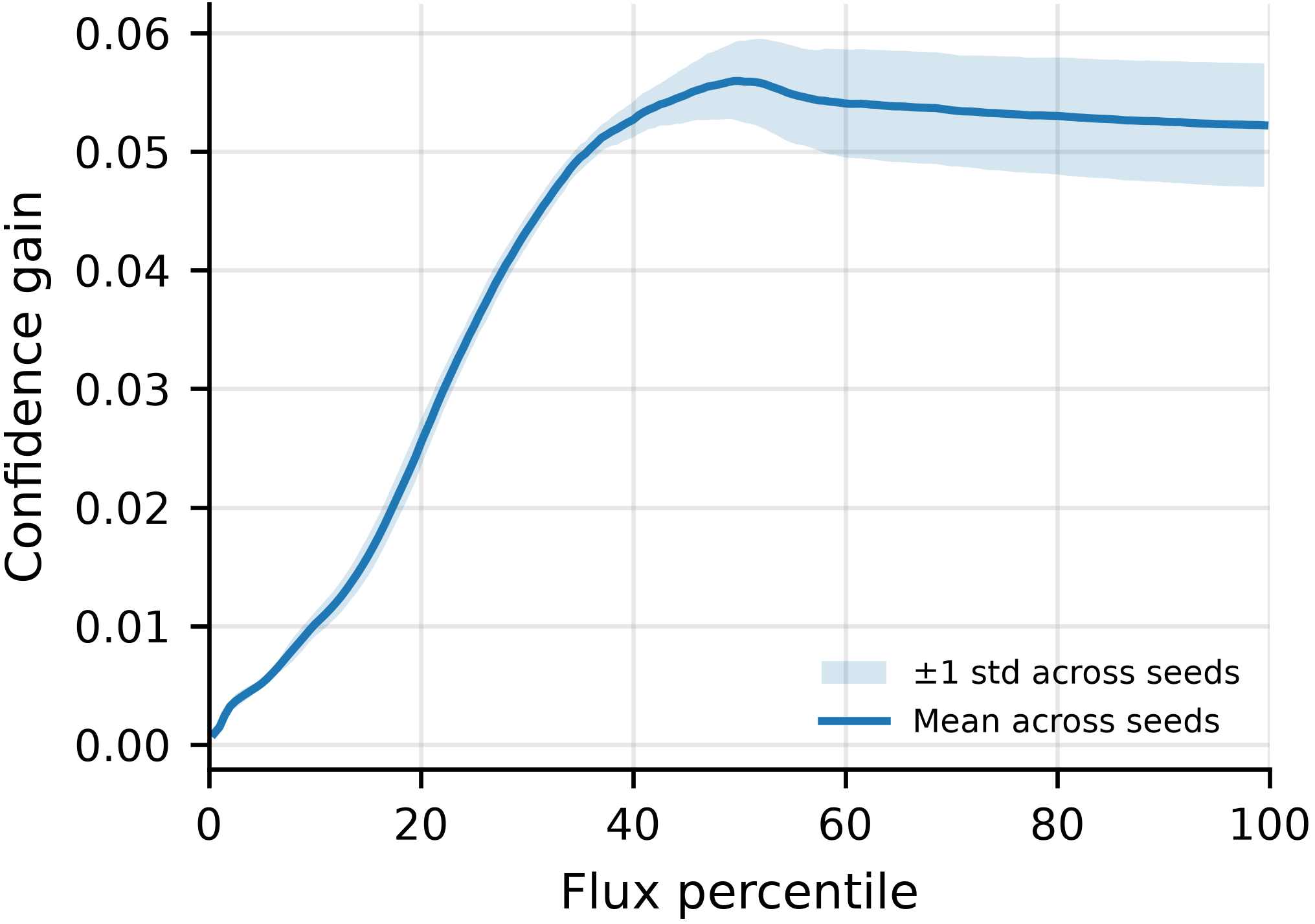}
    \end{subfigure}
    \hfill
    \begin{subfigure}[t]{0.32\textwidth}
        \centering
        \includegraphics[width=\linewidth]{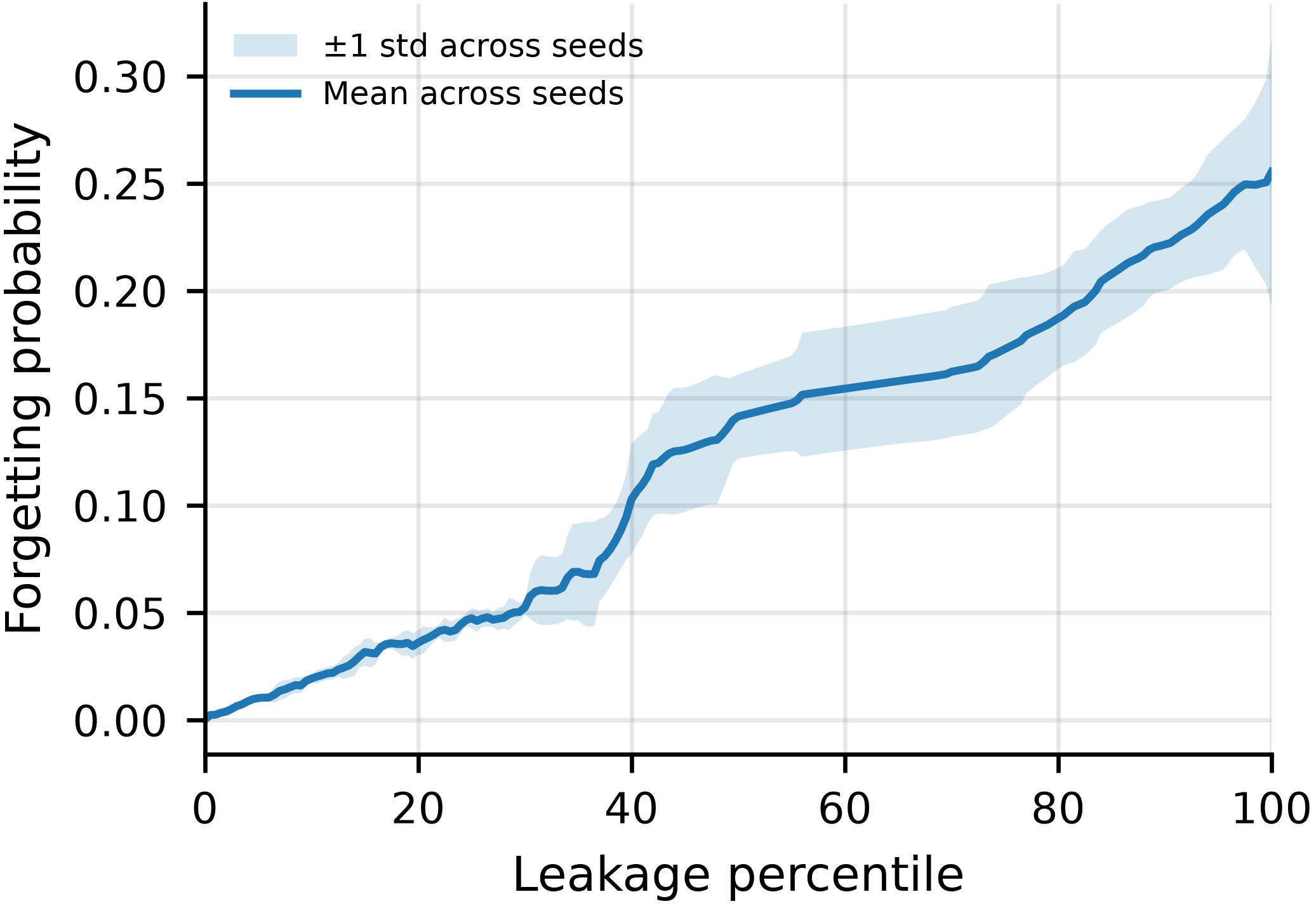}
    \end{subfigure}

    \vspace{3mm}

    \textbf{SplitTinyImageNet}\\[2mm]

    \begin{subfigure}[t]{0.32\textwidth}
        \centering
        \includegraphics[width=\linewidth]{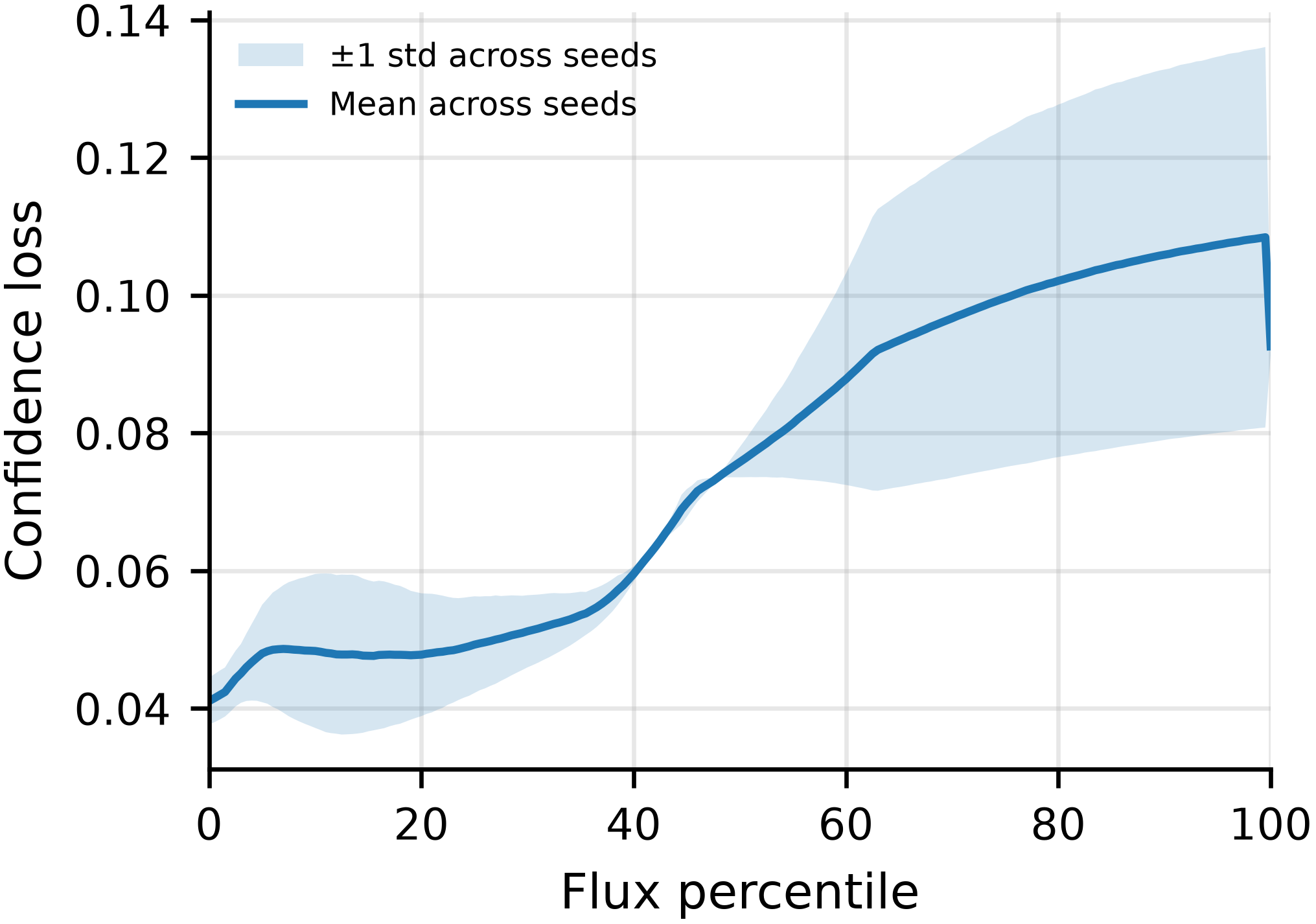}
    \end{subfigure}
    \hfill
    \begin{subfigure}[t]{0.32\textwidth}
        \centering
        \includegraphics[width=\linewidth]{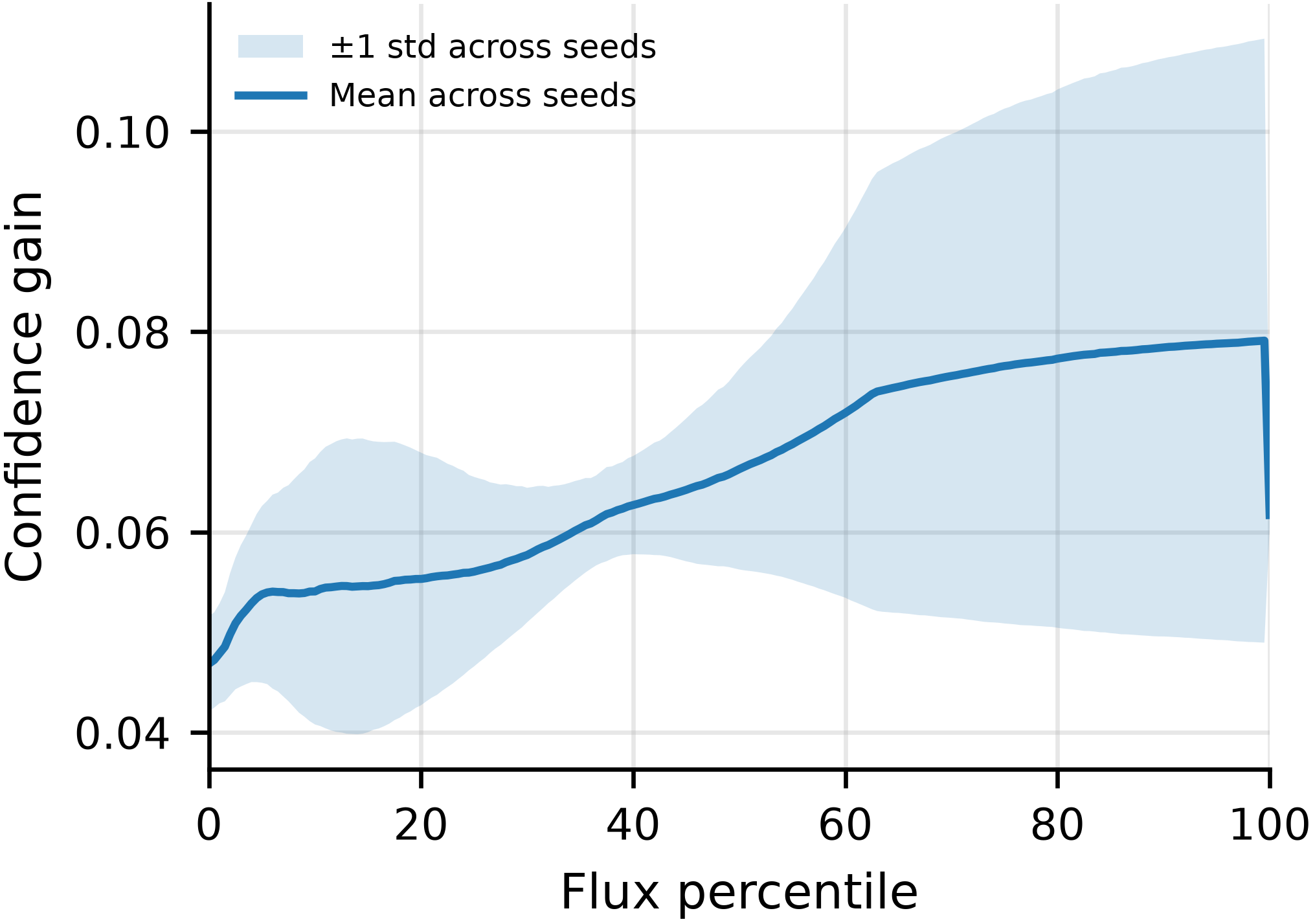}
    \end{subfigure}
    \hfill
    \begin{subfigure}[t]{0.32\textwidth}
        \centering
        \includegraphics[width=\linewidth]{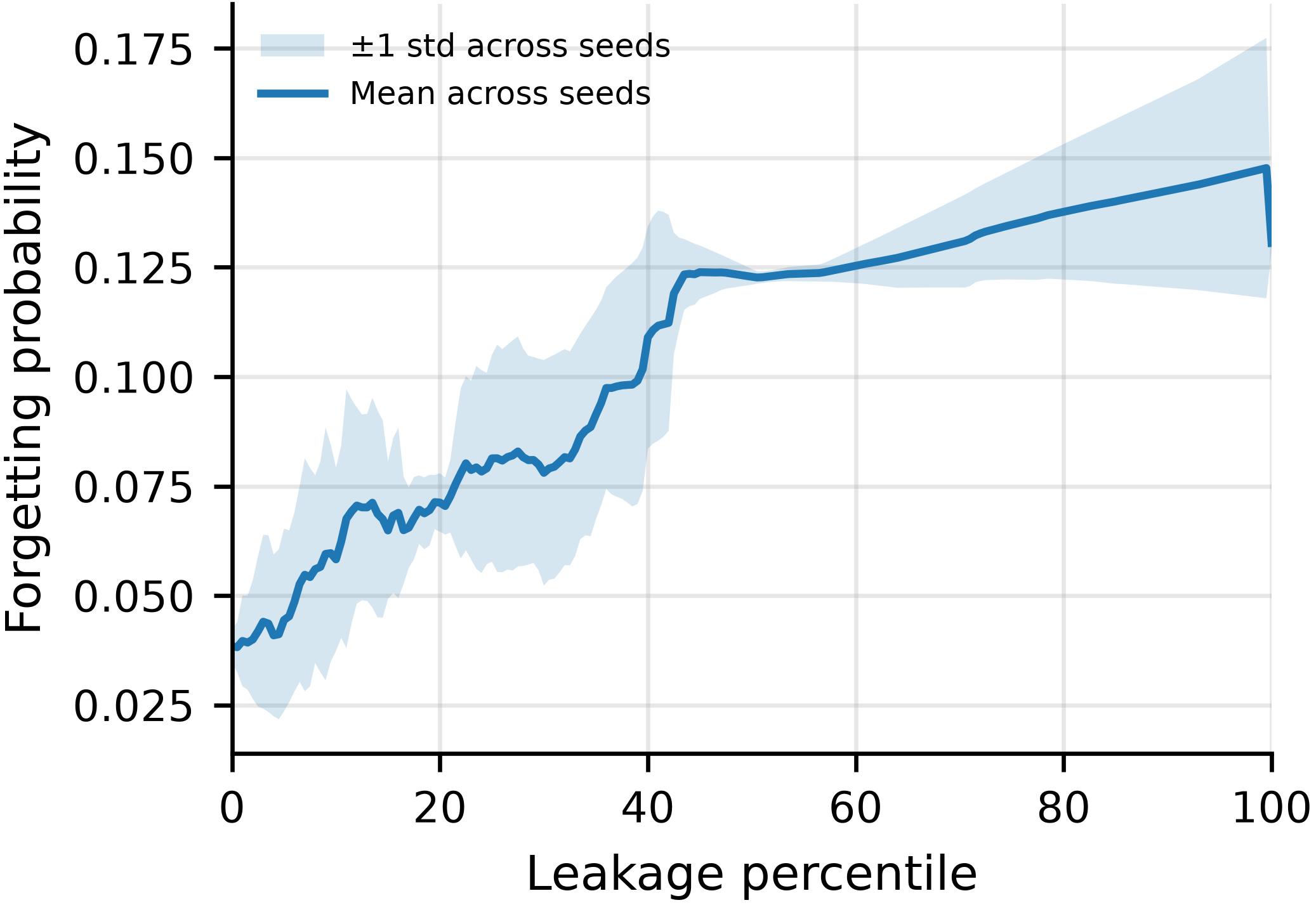}
    \end{subfigure}

    \caption{
    \textbf{Additional results on SplitFashionMNIST, SplitCIFAR10, and
SplitTinyImageNet.}
Across all datasets, representation flux is positively associated with
confidence loss, confidence gain is concentrated at substantially lower
levels of representation flux, and forgetting probability generally
increases with density leakage, with the largest forgetting probabilities
observed at high leakage values. 
}

    \label{fig:appendix_other_datasets}
\end{figure*}

Figure~\ref{fig:appendix_other_datasets} extends the geometric analysis to
SplitFashionMNIST, SplitCIFAR10, and SplitTinyImageNet. Each row
corresponds to one dataset, while the three columns report, respectively,
confidence loss as a function of representation flux, confidence gain as a
function of representation flux, and forgetting probability as a function
of density leakage.

Confidence loss denotes the decrease in the predicted probability assigned
to the ground-truth class,
\[
\Delta p^{-}
=
\max\!\left(
p_t(y\mid x)-p_{t+1}(y\mid x),\,0
\right),
\]
while confidence gain measures the corresponding increase,
\[
\Delta p^{+}
=
\max\!\left(
p_{t+1}(y\mid x)-p_t(y\mid x),\,0
\right).
\]

For visualization, samples are ordered according to representation flux,
and a centered rolling average is used to estimate the relationship between
flux and each outcome. For each random seed, observations outside the
$1^\mathrm{st}$ and $99^\mathrm{th}$ percentiles of the flux distribution
are excluded, and the remaining flux values are linearly rescaled to the
interval $[0,100]$. The resulting curves are interpolated onto a common
grid and averaged across seeds, with shaded regions indicating one standard
deviation across seeds. This normalization facilitates comparison across
runs and datasets despite differences in the absolute scale of their latent
representations.

The third column reports the forgetting probability,
\[
P(\mathrm{forgotten}\mid L),
\]
defined as the empirical probability that a sample transitions from correct
to incorrect classification between consecutive training snapshots, as a
function of density leakage. Density leakage is derived from the Density Transition Matrix
(Appendix~\ref{app:theory_stability}) and measures the probability that a
representation transitions from its initial latent-space region to a
different region between consecutive training snapshots. Larger leakage
values therefore indicate less stable occupancy of the discretized
latent-space regions.

\begin{figure*}[t]
    \centering

    \textbf{SplitFashionMNIST}\\[2mm]

    \begin{subfigure}[t]{0.48\textwidth}
        \centering
        \includegraphics[width=\linewidth]{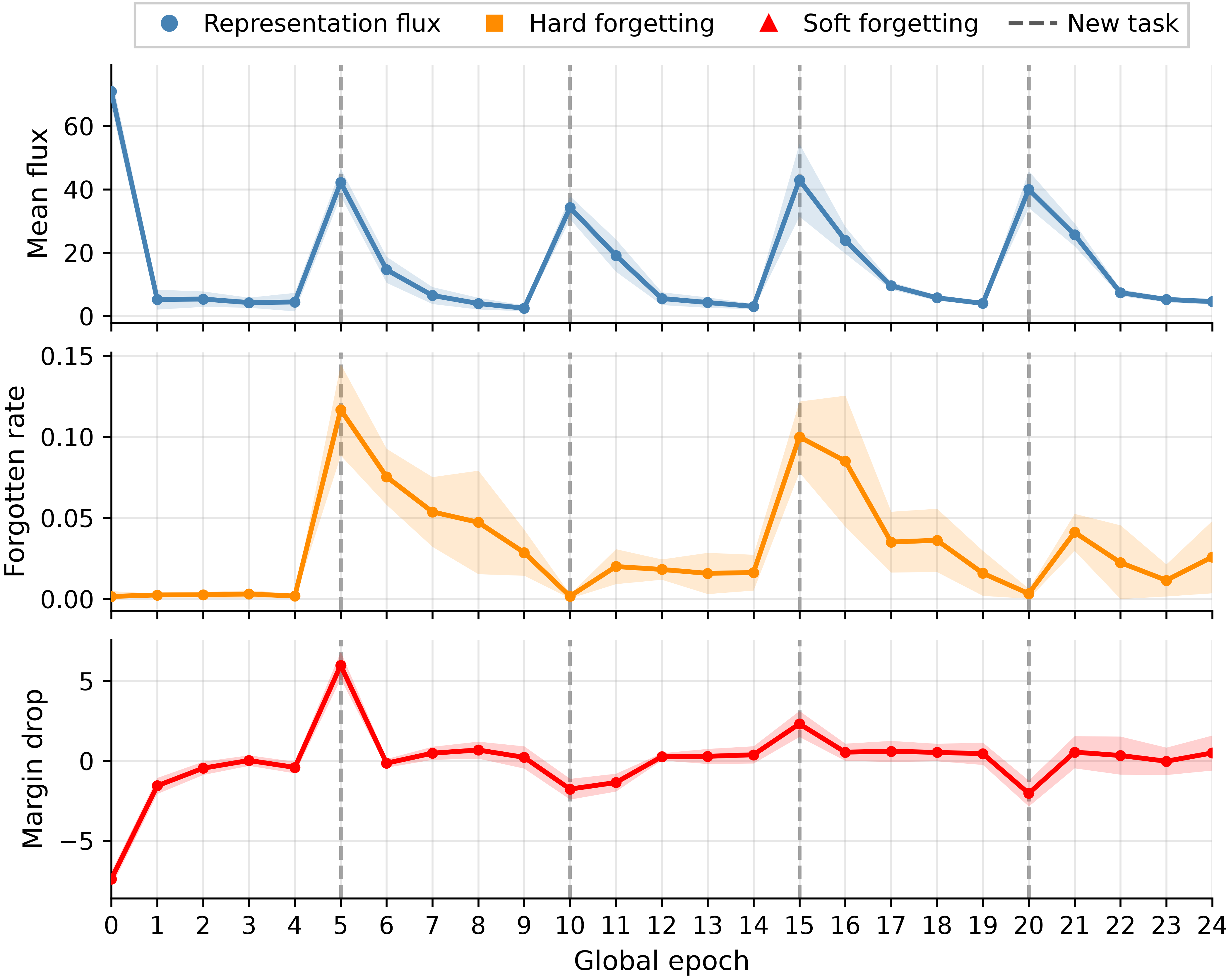}
    \end{subfigure}
    \hfill
    \begin{subfigure}[t]{0.48\textwidth}
        \centering
        \includegraphics[width=\linewidth]{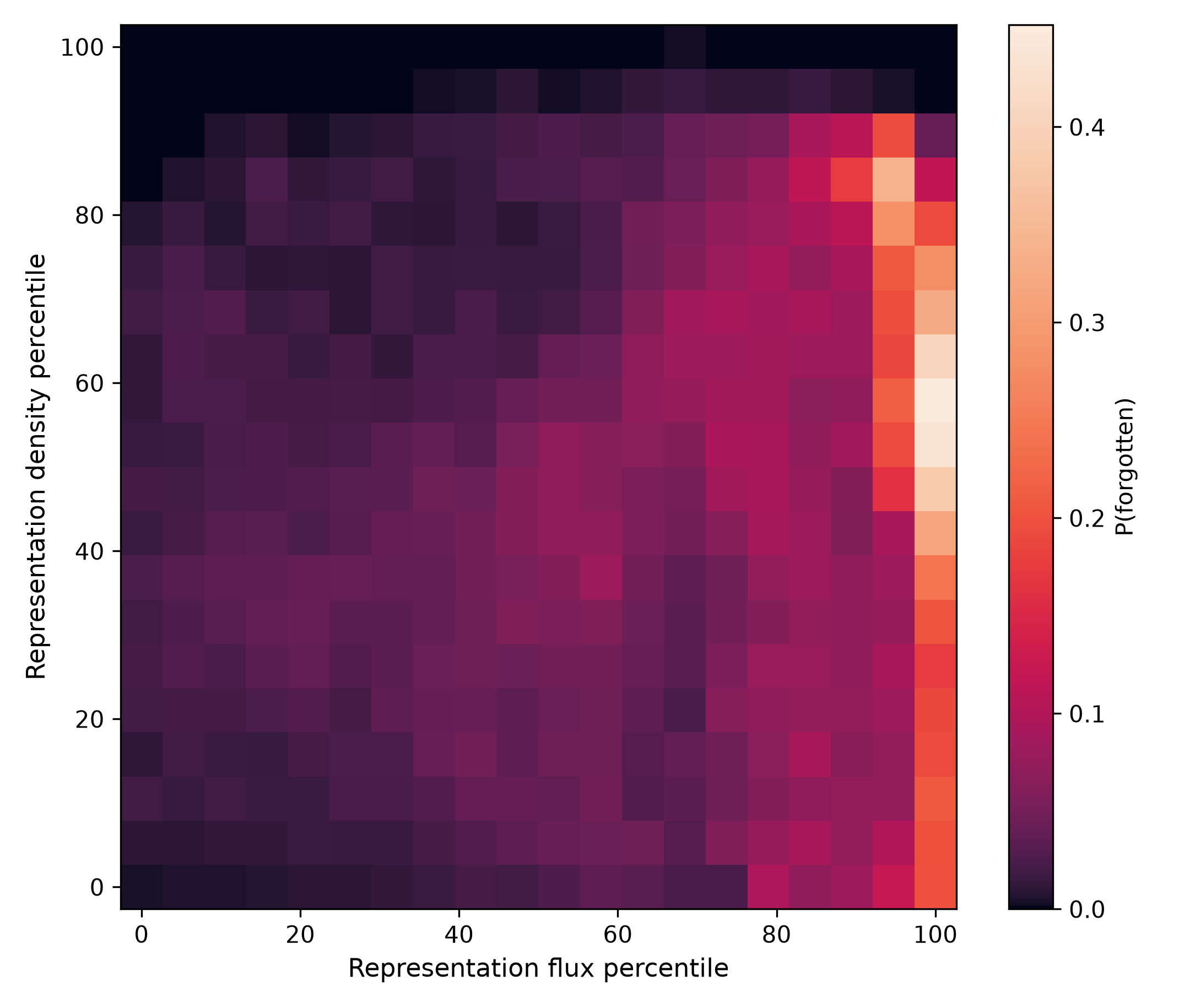}
    \end{subfigure}

    \vspace{3mm}

    \textbf{SplitCIFAR10}\\[2mm]

    \begin{subfigure}[t]{0.48\textwidth}
        \centering
        \includegraphics[width=\linewidth]{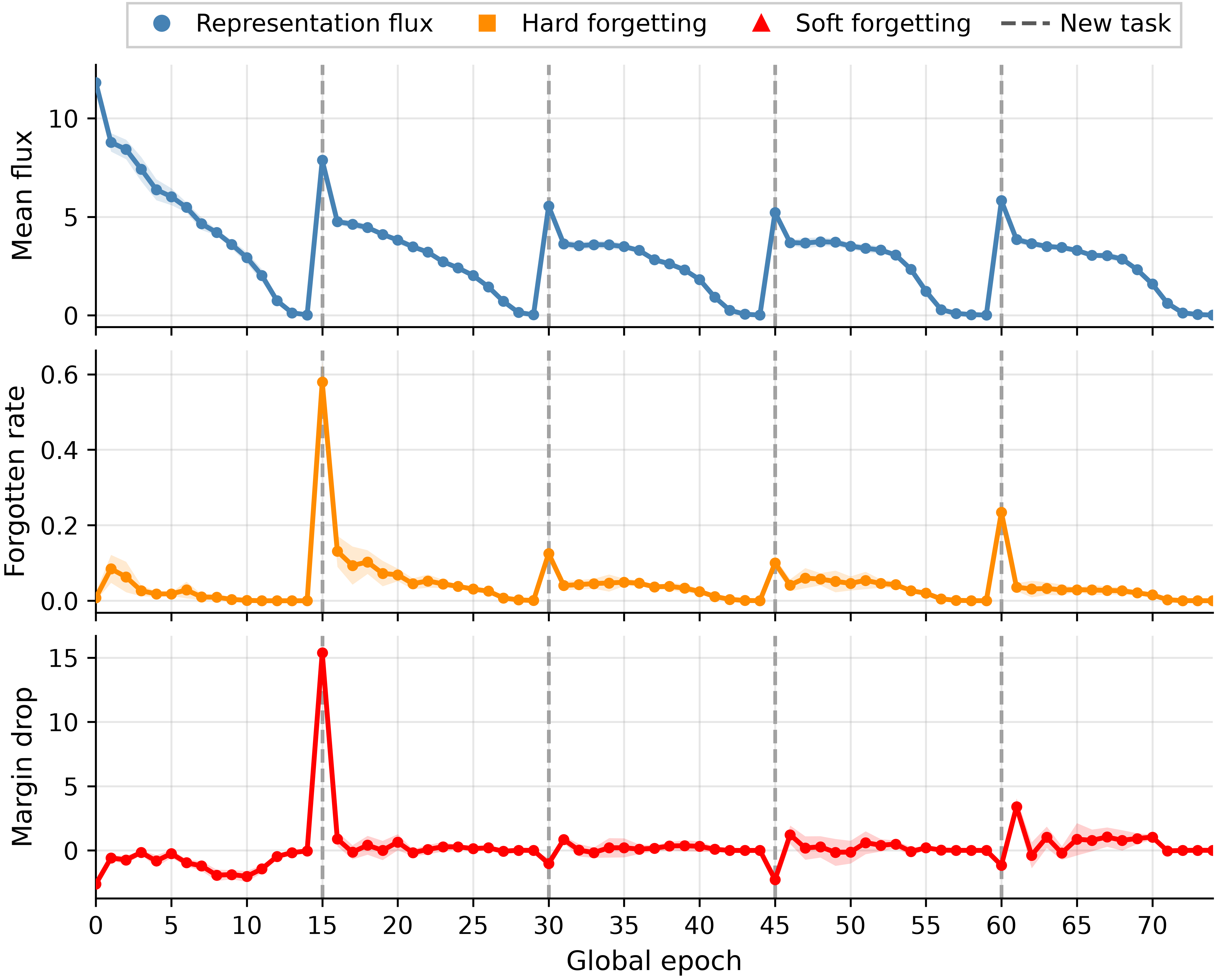}
    \end{subfigure}
    \hfill
    \begin{subfigure}[t]{0.48\textwidth}
        \centering
        \includegraphics[width=\linewidth]{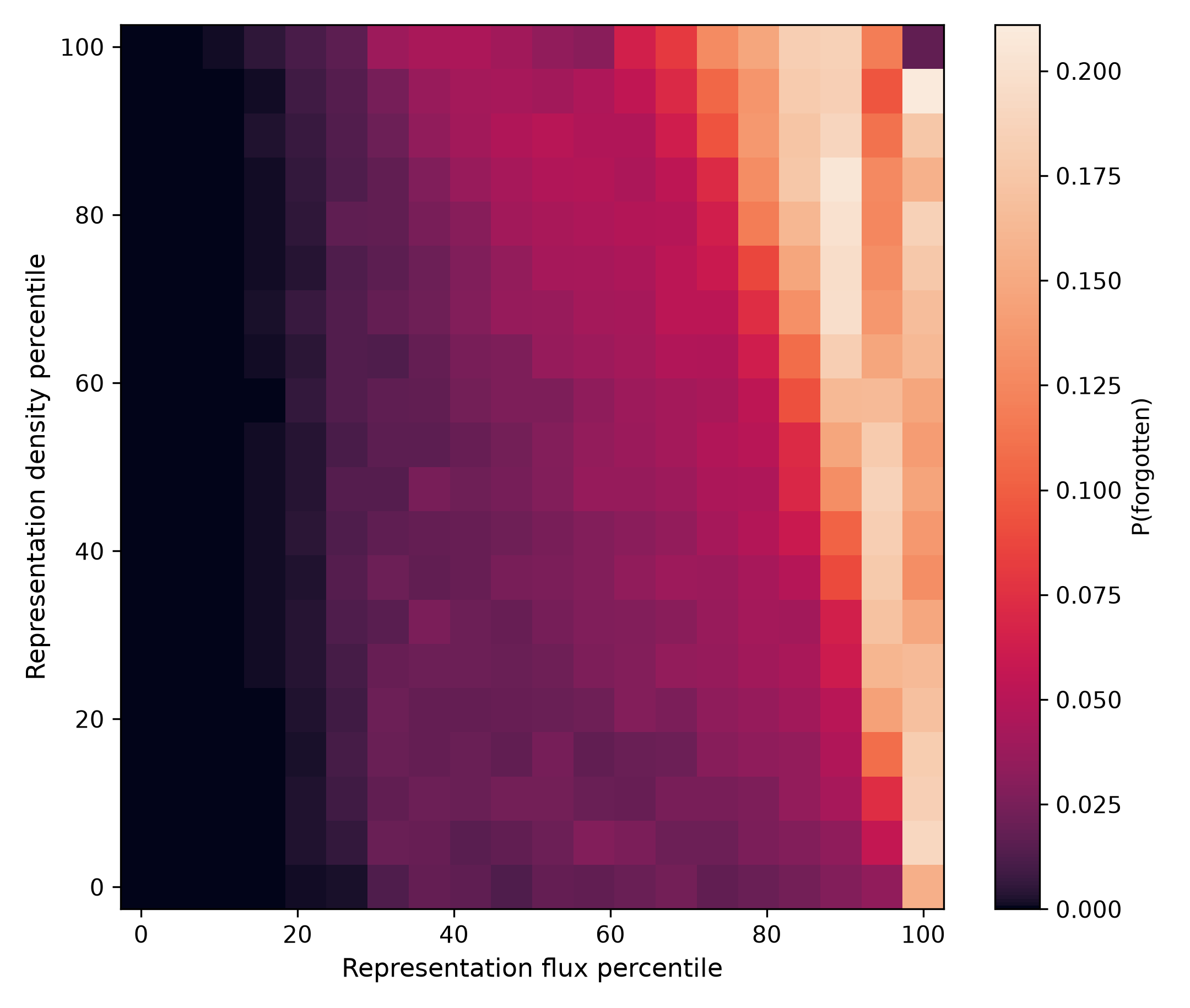}
    \end{subfigure}

    \vspace{3mm}

    \textbf{SplitTinyImageNet}\\[2mm]

    \begin{subfigure}[t]{0.48\textwidth}
        \centering
        \includegraphics[width=\linewidth]{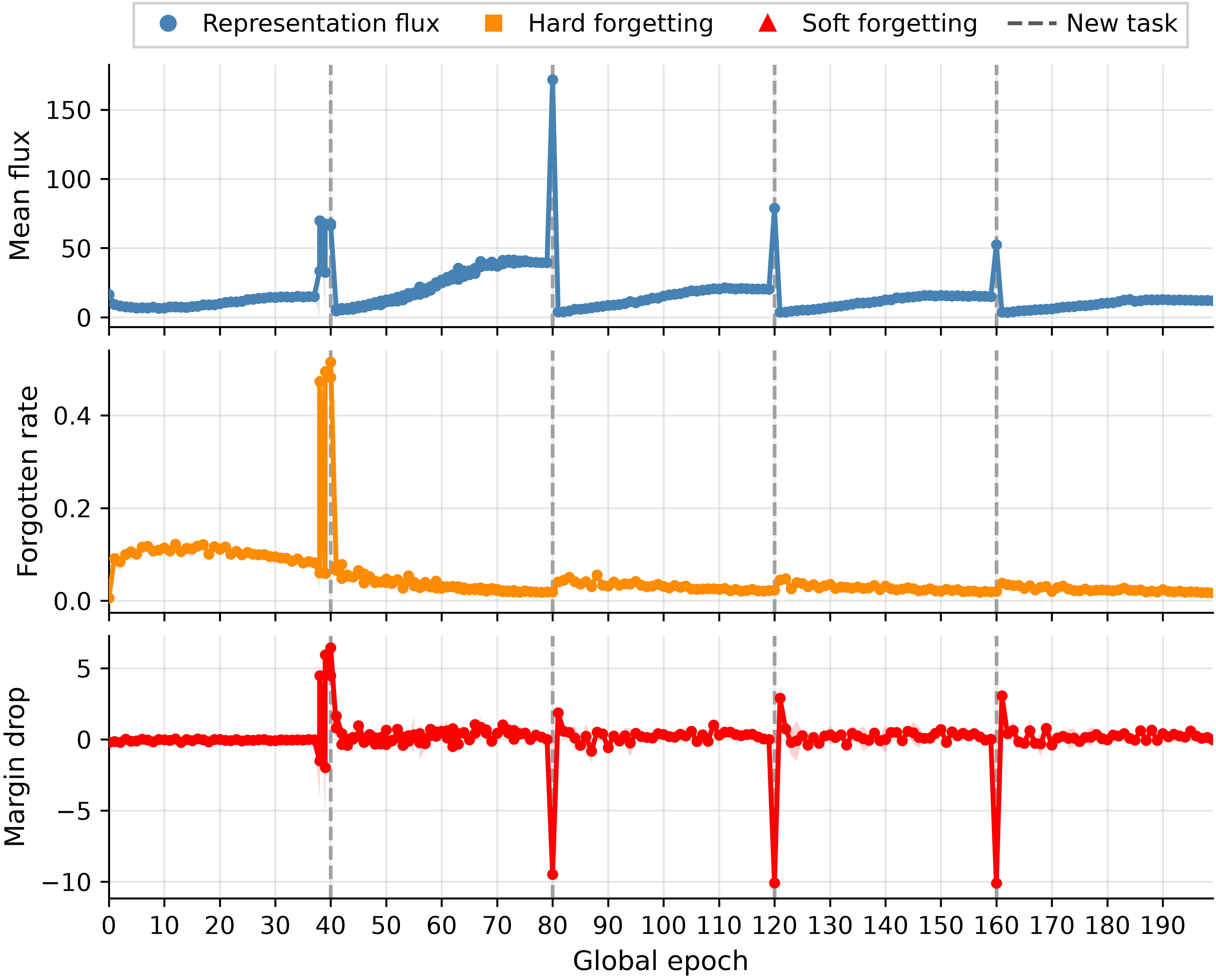}
    \end{subfigure}
    \hfill
    \begin{subfigure}[t]{0.48\textwidth}
        \centering
        \includegraphics[width=\linewidth]{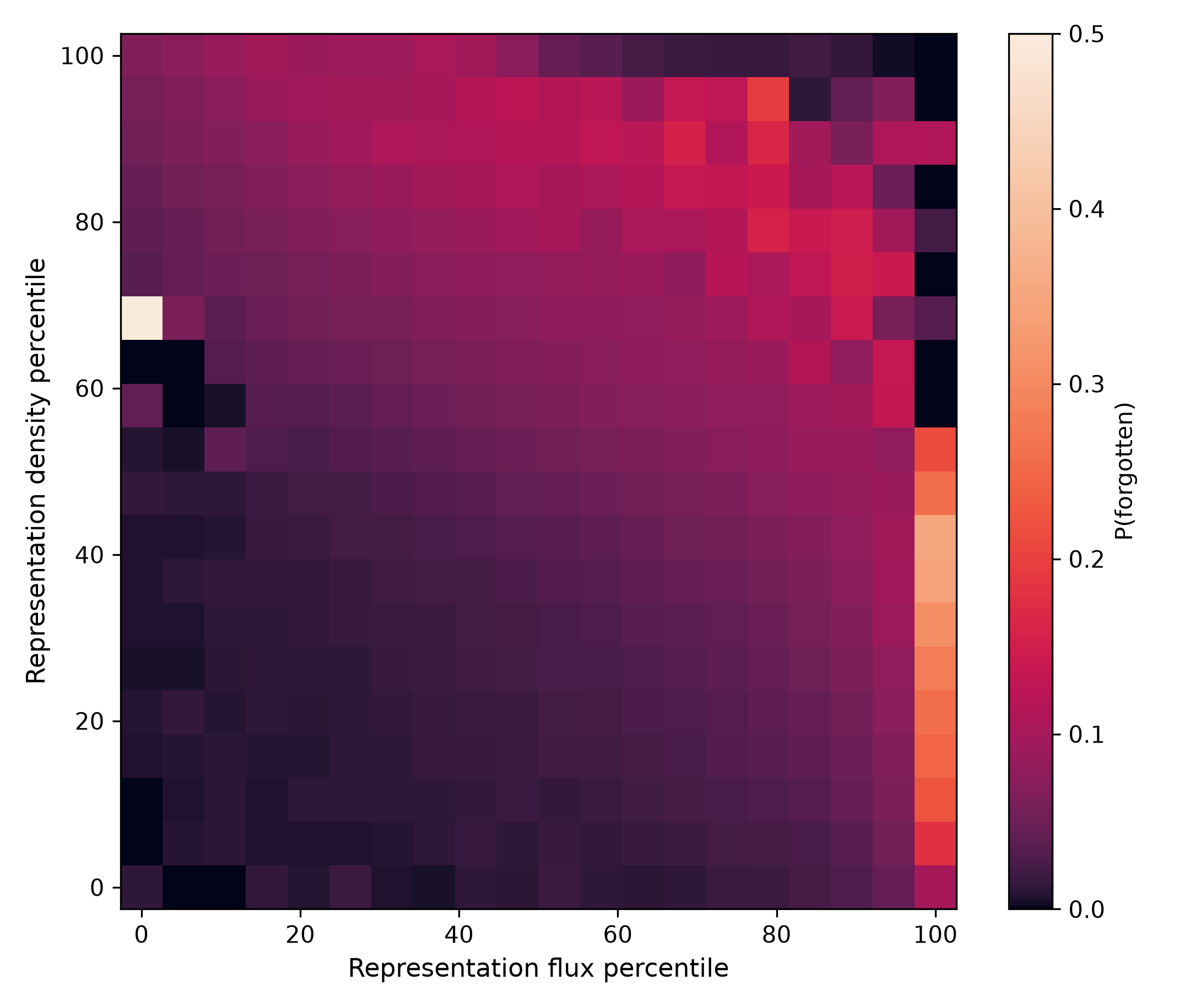}
    \end{subfigure}

    \caption{
\textbf{Geometric analyses on SplitFashionMNIST, SplitCIFAR10, and
SplitTinyImageNet.}
The left column shows the temporal evolution of representation flux
together with hard and soft forgetting across task transitions. The right
column presents forgetting probability as a function of representation
density and representation flux. Across datasets, elevated representation
flux is associated with subsequent forgetting, although the temporal
separation is less pronounced on the more complex benchmarks. High
representation flux combined with low representation density is associated
with the highest forgetting probability.
}

    \label{fig:appendix_flux_prediction}
\end{figure*}

Figure~\ref{fig:appendix_flux_prediction} complements these analyses by
examining the temporal evolution of representation dynamics during continual
learning. The left column contains three complementary temporal analyses of
representation flux and forgetting. The first compares mean representation
flux with the hard forgetting rate, defined as the fraction of previously
correct samples that become misclassified between consecutive training
snapshots. The second compares representation flux with soft forgetting,
measured as the decrease in predicted probability assigned to the
ground-truth class, $\Delta p^{-}$, providing a continuous measure of
degradation that does not require a change in predicted class. The third
compares representation flux with the mean margin drop, computed as
$m_t-m_{t+1}$, with positive values indicating a decrease in classification
margin (Appendix~\ref{app:theory_flux}). Together, these analyses characterize
the temporal relationship between representation displacement and different
measures of performance degradation.

The right column presents forgetting probability jointly as a function of
representation flux and representation density, with both quantities
converted to empirical percentile ranks for visualization. Representation
density is estimated from the distances to the $k=10$ nearest neighbors in
latent space, with higher values corresponding to samples located in more
densely populated regions. The resulting heatmaps therefore characterize
how representation displacement and local geometric structure are jointly
associated with forgetting.

The results indicate that several of the geometric relationships observed on
SplitMNIST remain qualitatively consistent across the additional benchmarks.
As shown in Figure~\ref{fig:appendix_other_datasets}, confidence loss
generally increases with representation flux, whereas confidence gain is
more concentrated within the low-flux regime. Thus, confidence
improvement is primarily associated with lower levels of representation
displacement, whereas larger displacement is predominantly associated with
confidence degradation. Forgetting probability also generally increases
with density leakage, indicating an association between instability of
latent-space region occupancy and forgetting.

The temporal analyses in Figure~\ref{fig:appendix_flux_prediction} show
that representation flux and forgetting exhibit related dynamics across the
additional datasets, although the temporal ordering in which elevated flux
precedes forgetting is less distinct than on SplitMNIST. The joint
density--flux heatmaps provide a complementary sample-level view: forgetting
is concentrated at higher representation flux, while local representation
density further modulates this relationship. This density dependence is
particularly pronounced on SplitTinyImageNet, where forgetting exhibits a
stronger concentration in lower-density regions than on the other evaluated
datasets. These results support representation flux as a recurring correlate
of forgetting across the evaluated benchmarks, while suggesting that local
representation density becomes increasingly informative on the more complex
SplitTinyImageNet benchmark.

\clearpage
\section{Experimental Setup}
\label{app:experimental_setup}

Unless otherwise stated, all experiments follow the standard
class-incremental learning (Class-IL) protocol
\citep{delange2021continual,vanDeVen2022three},
in which tasks are learned sequentially and the model predicts over all
classes observed so far without access to task identity at inference time.
Training data from previous tasks are unavailable except through the replay
buffer.

\paragraph{Datasets and architectures.}
SplitMNIST and SplitFashionMNIST use a fully connected multilayer perceptron
with an input dimension of 784, three hidden layers of 256, 256, and 64 units,
respectively, and a linear output layer. Each hidden layer is followed by a
ReLU activation. FlowLess-R regularization is applied to the 64-dimensional
representation produced by the third hidden layer immediately before the
classifier.

SplitCIFAR10 and SplitTinyImageNet use a standard ResNet-18 architecture
trained from scratch. The network consists of a convolutional stem followed
by four residual stages. Adaptive global average pooling is applied to the
output of the fourth residual stage to produce a 512-dimensional feature
representation, which is passed directly to a linear classification layer
with 10 output units for SplitCIFAR10 and 200 output units for
SplitTinyImageNet. FlowLess-R regularization is applied to this
512-dimensional representation immediately before the final classifier.

\paragraph{Experience replay.}
Experience replay (ER) uses a fixed per-task replay memory. After completing
training on each task, a fixed number of training samples is selected uniformly
at random without replacement and added to the replay buffer. For each selected
sample, the corresponding feature representation is extracted using the model
at the time of insertion and stored as a reference representation for
FlowLess-R. During subsequent tasks, replay samples are drawn uniformly without
replacement from the accumulated buffer at every optimization step, with replay
batches containing up to twice as many samples as the current-task mini-batch.
The replay classification loss is weighted by $2.0$ unless otherwise specified.

\paragraph{Optimization.}
SplitMNIST and SplitFashionMNIST models were trained for 5 epochs per task
using Adam with a learning rate of $10^{-3}$ and a mini-batch size of 256.
SplitCIFAR10 models were trained for 15 epochs per task using SGD with a
learning rate of $0.1$, momentum $0.9$, weight decay $5\times10^{-4}$, and a
mini-batch size of 128, with cosine annealing of the learning rate.
SplitTinyImageNet models were trained for 40 epochs per task using SGD with a
learning rate of $0.1$, momentum $0.9$, weight decay $5\times10^{-4}$, and a
mini-batch size of 64. For SplitTinyImageNet, a 5-epoch linear learning-rate
warm-up followed by cosine annealing was used.

\paragraph{FlowLess-R hyperparameters.}
For all experiments and all replay-based algorithms, the representation
regularization coefficient was selected from
\[
\lambda \in \{0,\,0.1,\,0.3,\,1.0,\,3.0\},
\]
where $\lambda=0$ corresponds to the baseline replay algorithm without
FlowLess-R.

\paragraph{ER buffer-size experiments.}
ER buffer-size experiments were conducted using memory sizes of
10, 20, 40, 80, and 160 samples per task for SplitMNIST and
SplitFashionMNIST; 50, 100, 200, 400, 800, and 1600 samples per task for
SplitCIFAR10; and 200, 400, 800, 1600, and 3200 samples per task for
SplitTinyImageNet. SplitMNIST, SplitFashionMNIST, and SplitCIFAR10 experiments
were repeated over ten random seeds (0--9), whereas SplitTinyImageNet
experiments were repeated over five random seeds (0--4).

After each task, the model was evaluated on the test sets of all tasks
encountered so far, yielding a task-by-task accuracy matrix. Final average
accuracy was computed as the average accuracy over all tasks after learning
the final task. Mean forgetting was computed as the average difference between
the maximum accuracy achieved on each previous task and its final accuracy
after learning all tasks. Unless otherwise stated, results are reported as
the mean $\pm$ standard deviation across the corresponding random seeds.

\paragraph{DER++ experiments.}
DER++ experiments were conducted using replay memories of 40 samples per task
for SplitMNIST and SplitFashionMNIST, 400 samples per task for SplitCIFAR10,
and 800 samples per task for SplitTinyImageNet. We implemented DER++ following
the original method of~\cite{buzzega2020dark}, with the addition
of the proposed FlowLess-R regularization term. The DER++ objective used a
logit-matching coefficient of $\alpha=0.1$ and a replay classification
coefficient of $\beta=2.0$. Gradient norms were clipped to $1.0$ for the
SplitCIFAR10 and SplitTinyImageNet experiments. All other experimental and
evaluation settings followed the corresponding ER experiments.

\paragraph{ER-ACE experiments.}
ER-ACE experiments were conducted using replay memories of 40 samples per task
for SplitMNIST and SplitFashionMNIST, 400 samples per task for SplitCIFAR10,
and 800 samples per task for SplitTinyImageNet. We implemented ER-ACE following
the original method of~\cite{caccia2022new}, with the addition of
the proposed FlowLess-R regularization term. For current-task samples, the
classification loss was computed after masking logits corresponding to classes
not observed up to and including the current task, whereas replay samples were
optimized using the standard unmasked cross-entropy loss. All other experimental
and evaluation settings followed the corresponding ER experiments.

\clearpage
\section{Additional FlowLess-R Results}
\label{app:add_flow}

\subsection{Final Average Accuracy}

Table~\ref{tab:accuracy} reports final average accuracy across all evaluated
values of the FlowLess-R regularization coefficient $\lambda$. FlowLess-R
generally maintains or improves final predictive performance relative to the
corresponding replay baseline. For ER, the largest improvements are observed
on SplitMNIST and SplitFashionMNIST, with gains of $6.81$ and $6.12$
percentage points, respectively ($p<0.0001$ in both cases). Smaller gains are
observed on SplitCIFAR10 ($+1.76$ points, $p=0.3389$) and
SplitTinyImageNet ($+2.43$ points, $p=0.1313$), neither of which reaches
statistical significance after multiple-comparison correction.

For DER++, substantial and statistically significant improvements are
observed on both ResNet-18 benchmarks. Final average accuracy increases from
$43.15\%$ to $49.69\%$ on SplitCIFAR10 ($+6.54$ points, $p=0.0082$) and
from $13.22\%$ to $18.78\%$ on SplitTinyImageNet ($+5.56$ points,
$p=0.0084$). Improvements on SplitMNIST ($+1.02$ points, $p=0.1133$) and
SplitFashionMNIST ($+0.59$ points, $p=0.1418$) are smaller and do not reach
statistical significance.

ER-ACE exhibits significant accuracy gains on SplitMNIST ($+4.87$ points,
$p<0.0001$) and SplitFashionMNIST ($+5.05$ points, $p<0.0001$), while the
improvements on SplitCIFAR10 ($+1.90$ points, $p=0.3309$) and
SplitTinyImageNet ($+1.91$ points, $p=0.0567$) do not reach statistical
significance after multiple-comparison correction.

Together with the forgetting results reported in the main text, these results
indicate that the improved retention produced by FlowLess-R is generally not
obtained at the expense of final predictive performance. In most evaluated
settings, FlowLess-R simultaneously reduces forgetting and improves final
average accuracy. In particular, the substantial gains for DER++ on
SplitCIFAR10 and SplitTinyImageNet show that stronger retention can coincide
with improved final predictive performance on the more visually complex
benchmarks.
\begin{table*}[t]
\centering
\caption{Final average accuracy (\%, mean $\pm$ standard deviation) for
different values of the FlowLess-R regularization coefficient $\lambda$.
Results are averaged over 10 random seeds for SplitMNIST,
SplitFashionMNIST, and SplitCIFAR10, and 5 random seeds for
SplitTinyImageNet. The setting $\lambda=0$ corresponds to the baseline
method without FlowLess-R regularization. Reported $p$-values are
Holm--Bonferroni corrected for multiple comparisons across $\lambda$
values.}
\label{tab:accuracy}
\resizebox{\textwidth}{!}{
\begin{tabular}{llcccc}
\toprule
Algorithm & $\lambda$ &
SplitMNIST &
SplitFashionMNIST &
SplitCIFAR10 &
SplitTinyImageNet \\
\midrule
\textbf{Backbone} & &
\textbf{MLP} &
\textbf{MLP} &
\textbf{ResNet-18} &
\textbf{ResNet-18} \\
\midrule
\textbf{Buffer / task} & &
\textbf{40} &
\textbf{40} &
\textbf{400} &
\textbf{800} \\
\midrule

\multirow{6}{*}{ER}
& 0.0 & 72.98 $\pm$ 1.45 & 64.73 $\pm$ 0.92 & 37.83 $\pm$ 2.75 & 16.33 $\pm$ 0.94 \\
\cmidrule(lr){2-6}
& 0.1 & 76.04 $\pm$ 1.27 & 67.63 $\pm$ 1.64 & 38.83 $\pm$ 3.61 & \textbf{18.76 $\pm$ 1.56} \\
& 0.3 & 78.98 $\pm$ 1.82 & 69.61 $\pm$ 0.82 & 38.92 $\pm$ 3.90 & 17.78 $\pm$ 1.33 \\
& 1.0 & \textbf{79.79 $\pm$ 1.25} & \textbf{70.85 $\pm$ 1.02} & \textbf{39.59 $\pm$ 2.31} & 15.94 $\pm$ 1.61 \\
& 3.0 & 78.38 $\pm$ 1.00 & 69.56 $\pm$ 1.67 & 39.27 $\pm$ 3.15 & 18.11 $\pm$ 1.70 \\
\cmidrule(lr){2-6}
& \textbf{Best $\Delta$} &
\textbf{+6.81 ($p<0.0001$)} &
\textbf{+6.12 ($p<0.0001$)} &
\textbf{+1.76 ($p=0.3389$)} &
\textbf{+2.43 ($p=0.1313$)} \\
\cmidrule(lr){2-6}

\multirow{6}{*}{DER++}
& 0.0 & 77.71 $\pm$ 1.84 & 70.91 $\pm$ 1.78 & 43.15 $\pm$ 4.73 & 13.22 $\pm$ 2.01 \\
\cmidrule(lr){2-6}
& 0.1 & 78.18 $\pm$ 1.77 & 70.93 $\pm$ 1.75 & 48.79 $\pm$ 0.83 & 15.89 $\pm$ 0.41 \\
& 0.3 & 77.64 $\pm$ 1.32 & 71.40 $\pm$ 1.63 & 48.26 $\pm$ 1.52 & 17.30 $\pm$ 0.75 \\
& 1.0 & 78.30 $\pm$ 0.87 & \textbf{71.50 $\pm$ 1.79} & 49.25 $\pm$ 1.96 & 18.10 $\pm$ 0.74 \\
& 3.0 & \textbf{78.74 $\pm$ 1.46} & 71.35 $\pm$ 1.84 & \textbf{49.69 $\pm$ 1.40} & \textbf{18.78 $\pm$ 0.62} \\
\cmidrule(lr){2-6}
& \textbf{Best $\Delta$}
& \textbf{+1.02 ($p=0.1133$)}
& \textbf{+0.59 ($p=0.1418$)}
& \textbf{+6.54 ($p=0.0082$)}
& \textbf{+5.56 ($p=0.0084$)} \\

\multirow{6}{*}{ER-ACE}
& 0.0 & 72.87 $\pm$ 2.03 & 64.12 $\pm$ 1.65 & 40.05 $\pm$ 3.54 & 16.43 $\pm$ 0.82 \\
\cmidrule(lr){2-6}
& 0.1 & 75.44 $\pm$ 1.32 & 65.65 $\pm$ 1.87 & 40.87 $\pm$ 1.01 & 18.14 $\pm$ 1.20 \\
& 0.3 & 77.09 $\pm$ 1.18 & 67.85 $\pm$ 1.59 & 41.37 $\pm$ 2.41 & 17.62 $\pm$ 1.50 \\
& 1.0 & \textbf{77.74 $\pm$ 1.04} & \textbf{69.17 $\pm$ 1.07} & \textbf{41.95 $\pm$ 1.84} & 14.93 $\pm$ 0.44 \\
& 3.0 & 77.45 $\pm$ 1.45 & 69.15 $\pm$ 1.17 & 40.57 $\pm$ 3.15 & \textbf{18.34 $\pm$ 0.91} \\
\cmidrule(lr){2-6}
& \textbf{Best $\Delta$}
& \textbf{+4.87 ($p<0.0001$)}
& \textbf{+5.05 ($p<0.0001$)}
& \textbf{+1.90 ($p=0.3309$)}
& \textbf{+1.91 ($p=0.0567$)} \\
\bottomrule
\end{tabular}}
\end{table*}
\subsection{Effect of Replay Buffer Size}
\label{sec:buffer_size}

To evaluate the robustness of FlowLess-R across different replay capacities, we varied the
replay buffer size over dataset-specific ranges and compared the resulting performance with
the corresponding baseline replay methods. For each replay buffer size, we selected the regularization coefficient $\lambda$
that achieved the highest final average accuracy on the validation runs, while the baseline
corresponds to $\lambda=0$. Figure~5 shows that FlowLess-R generally improves continual
learning performance across replay budgets. On SplitMNIST and SplitFashionMNIST, the gains
are largest at smaller replay capacities and gradually narrow as the buffer grows, consistent
with diminishing benefits from additional representation stabilization when replay itself
becomes increasingly effective. SplitCIFAR10 exhibits a less pronounced and less monotonic
improvement, with substantial variability in forgetting across seeds.

In contrast, the effect is particularly pronounced on SplitTinyImageNet at larger replay
capacities: the performance gap widens substantially as the buffer increases, accompanied
by a marked reduction in forgetting. One possible explanation is that, for the more complex
and diverse TinyImageNet representation space, larger buffers provide FlowLess-R with a
broader set of stored representation anchors, allowing the regularizer to constrain
representation drift more effectively. On SplitCIFAR10, the smaller number of classes and
lower visual diversity may make additional replay samples increasingly redundant, reducing
the incremental benefit of representation stabilization. These interpretations are
speculative, however, and the buffer-size experiments do not directly identify the mechanism
responsible for the different scaling behavior across datasets.
\begin{figure*}[!ht]
    \centering

    \begin{subfigure}[t]{0.24\textwidth}
        \centering
        \includegraphics[width=\linewidth]{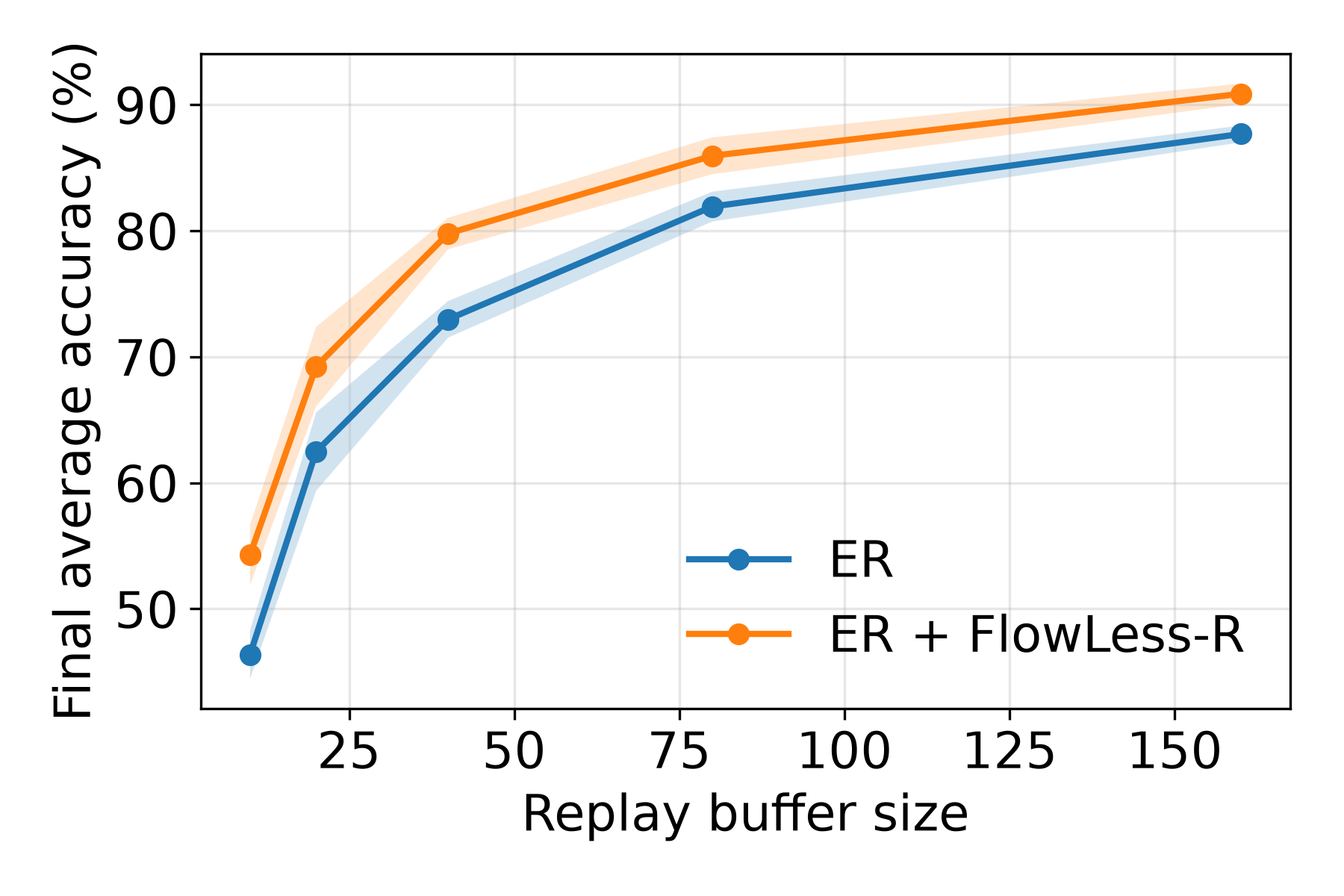}
        \caption{SplitMNIST}
    \end{subfigure}
    \hfill
    \begin{subfigure}[t]{0.24\textwidth}
        \centering
        \includegraphics[width=\linewidth]{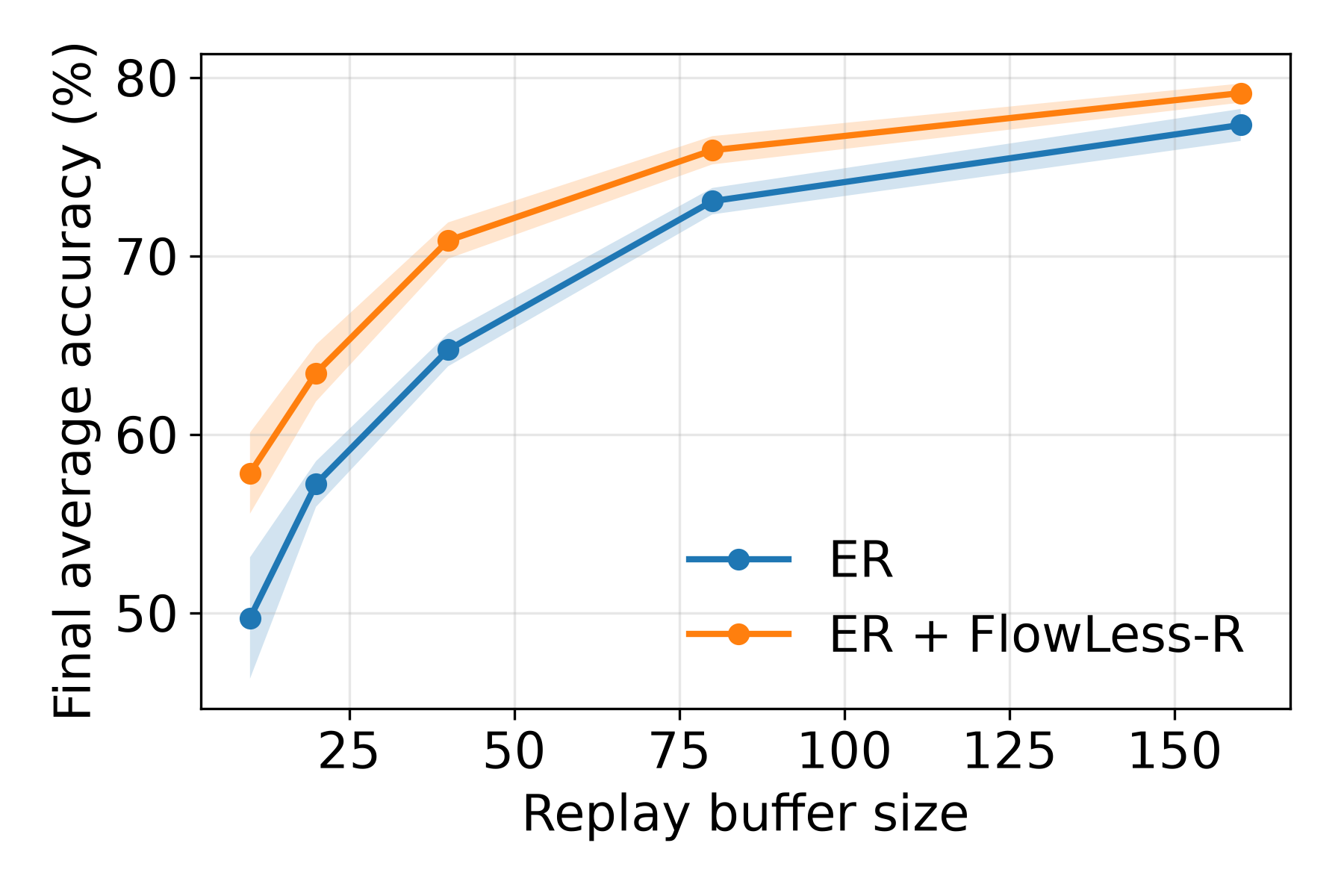}
        \caption{SplitFashionMNIST}
    \end{subfigure}
    \hfill
    \begin{subfigure}[t]{0.24\textwidth}
        \centering
        \includegraphics[width=\linewidth]{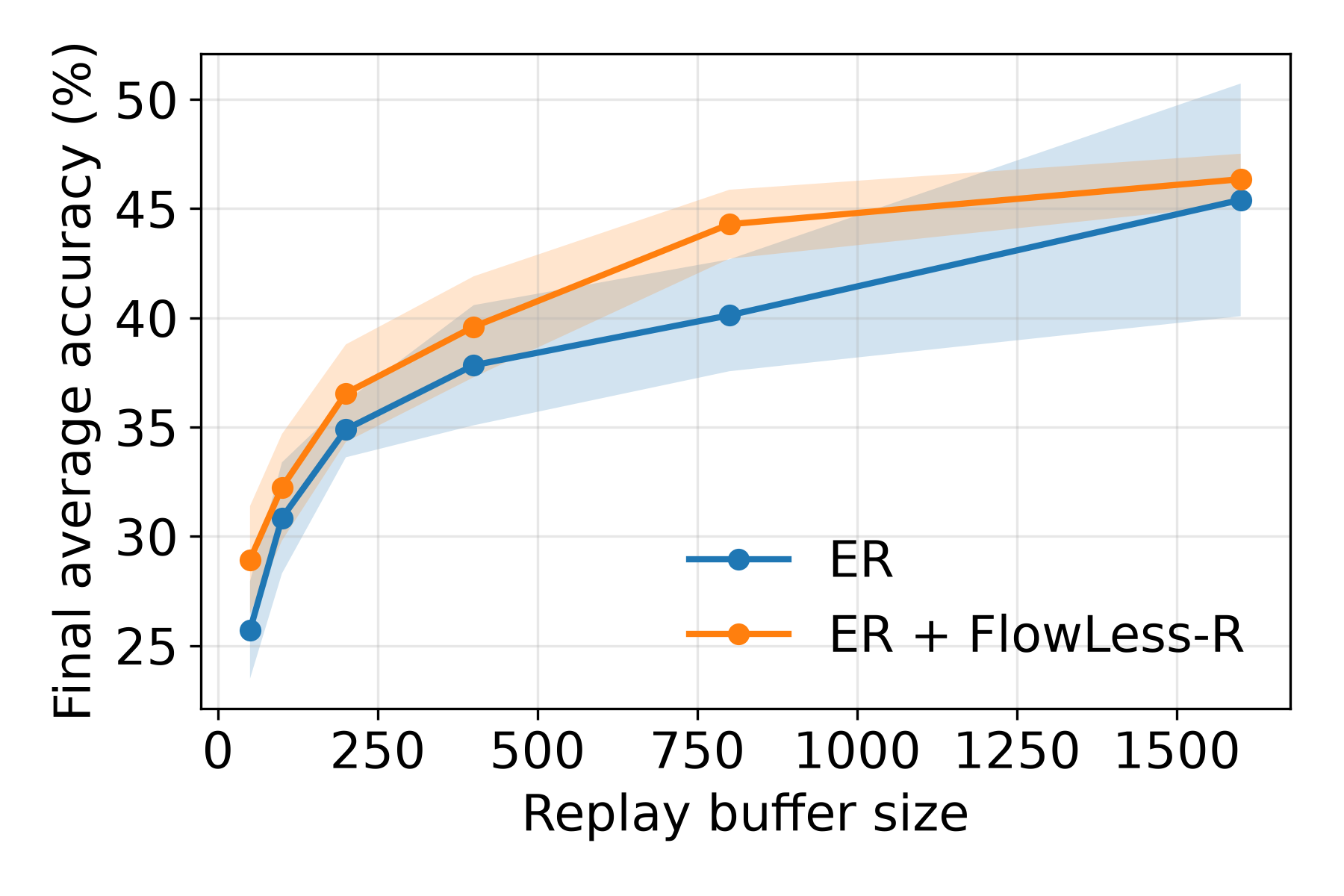}
        \caption{SplitCIFAR10}
    \end{subfigure}
    \hfill
    \begin{subfigure}[t]{0.24\textwidth}
        \centering
        \includegraphics[width=\linewidth]{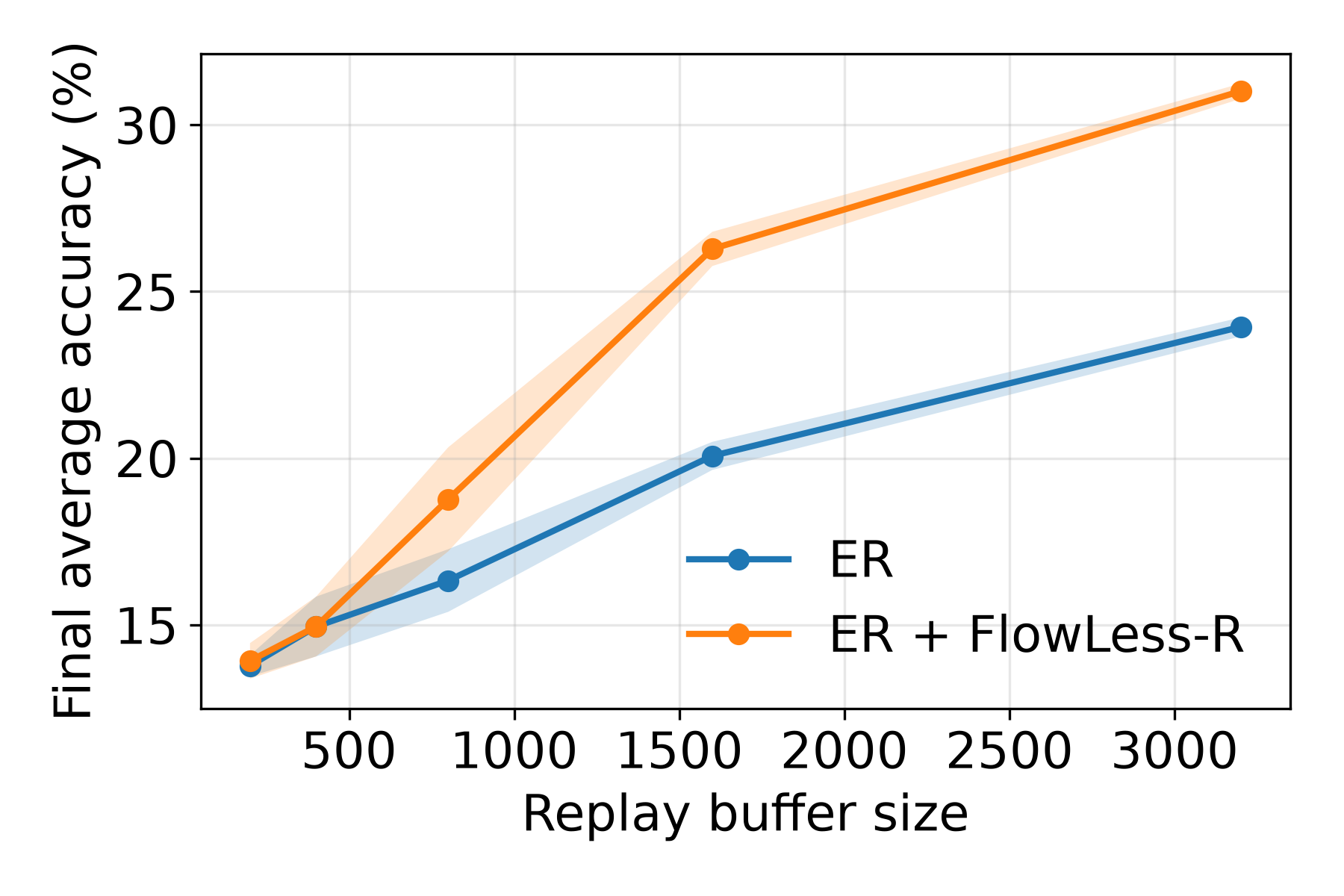}
        \caption{SplitTinyImageNet}
    \end{subfigure}

    \vspace{0.4cm}

    \begin{subfigure}[t]{0.24\textwidth}
        \centering
        \includegraphics[width=\linewidth]{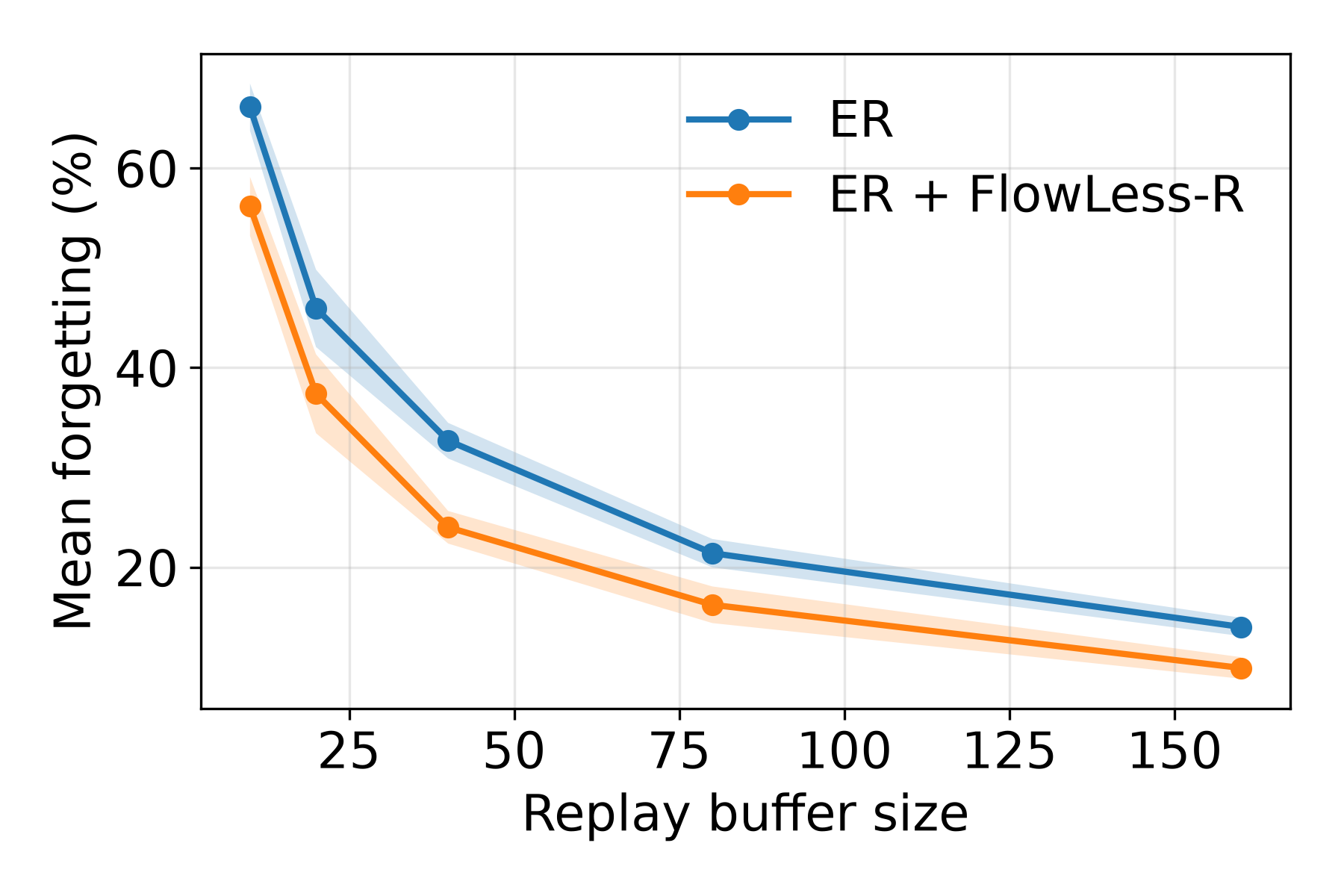}
        \caption{SplitMNIST}
    \end{subfigure}
    \hfill
    \begin{subfigure}[t]{0.24\textwidth}
        \centering
        \includegraphics[width=\linewidth]{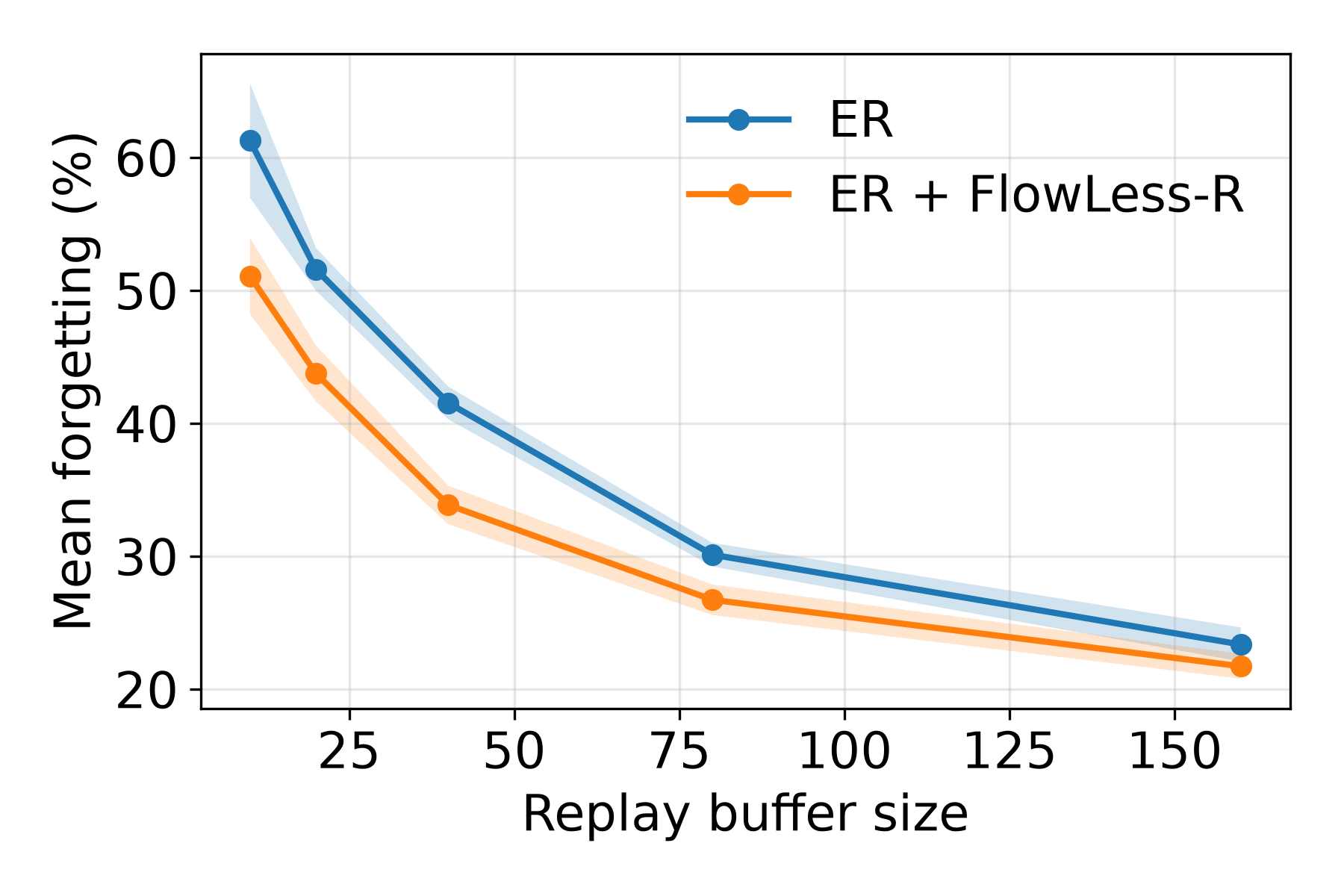}
        \caption{SplitFashionMNIST}
    \end{subfigure}
    \hfill
    \begin{subfigure}[t]{0.24\textwidth}
        \centering
        \includegraphics[width=\linewidth]{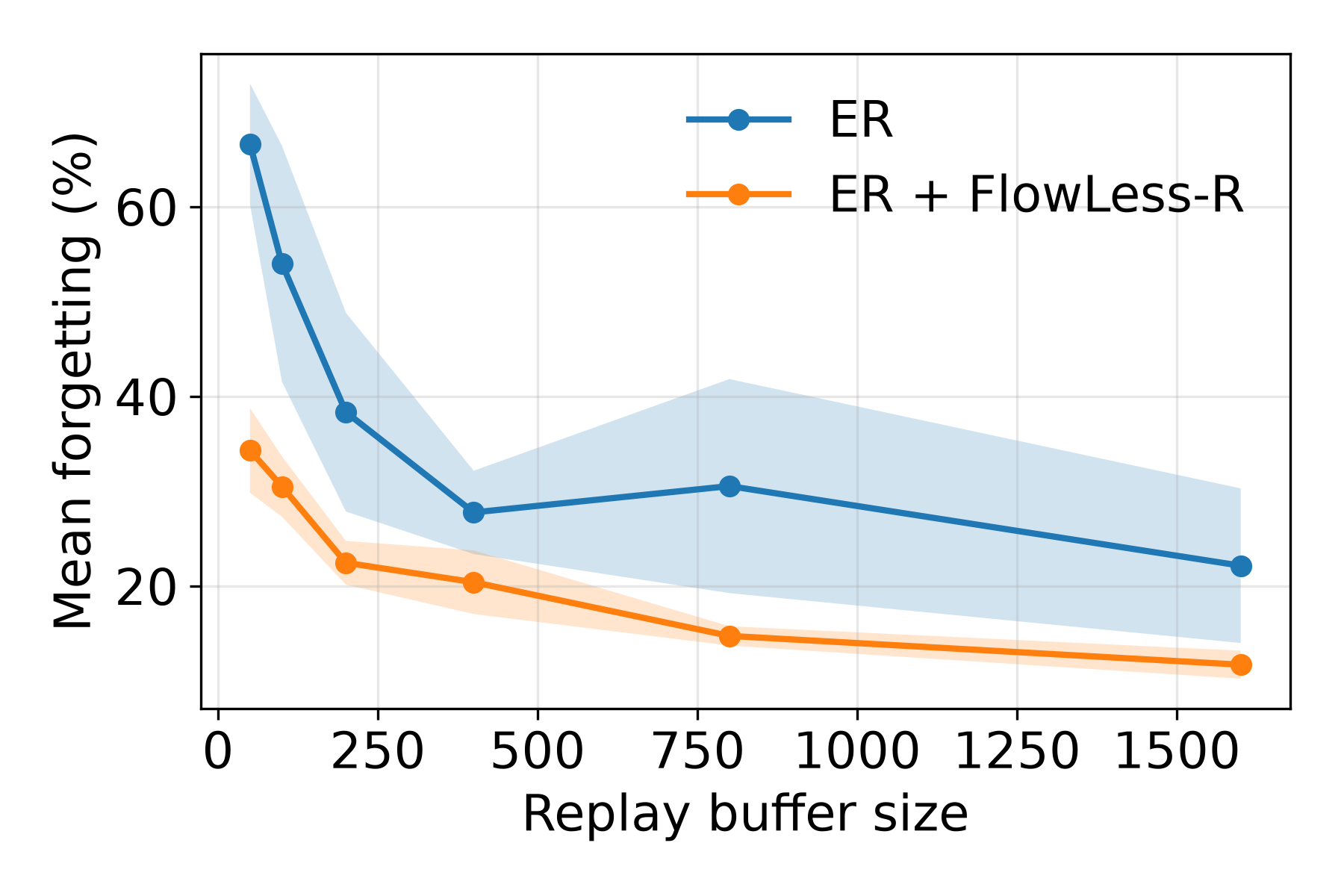}
        \caption{SplitCIFAR10}
    \end{subfigure}
    \hfill
    \begin{subfigure}[t]{0.24\textwidth}
        \centering
        \includegraphics[width=\linewidth]{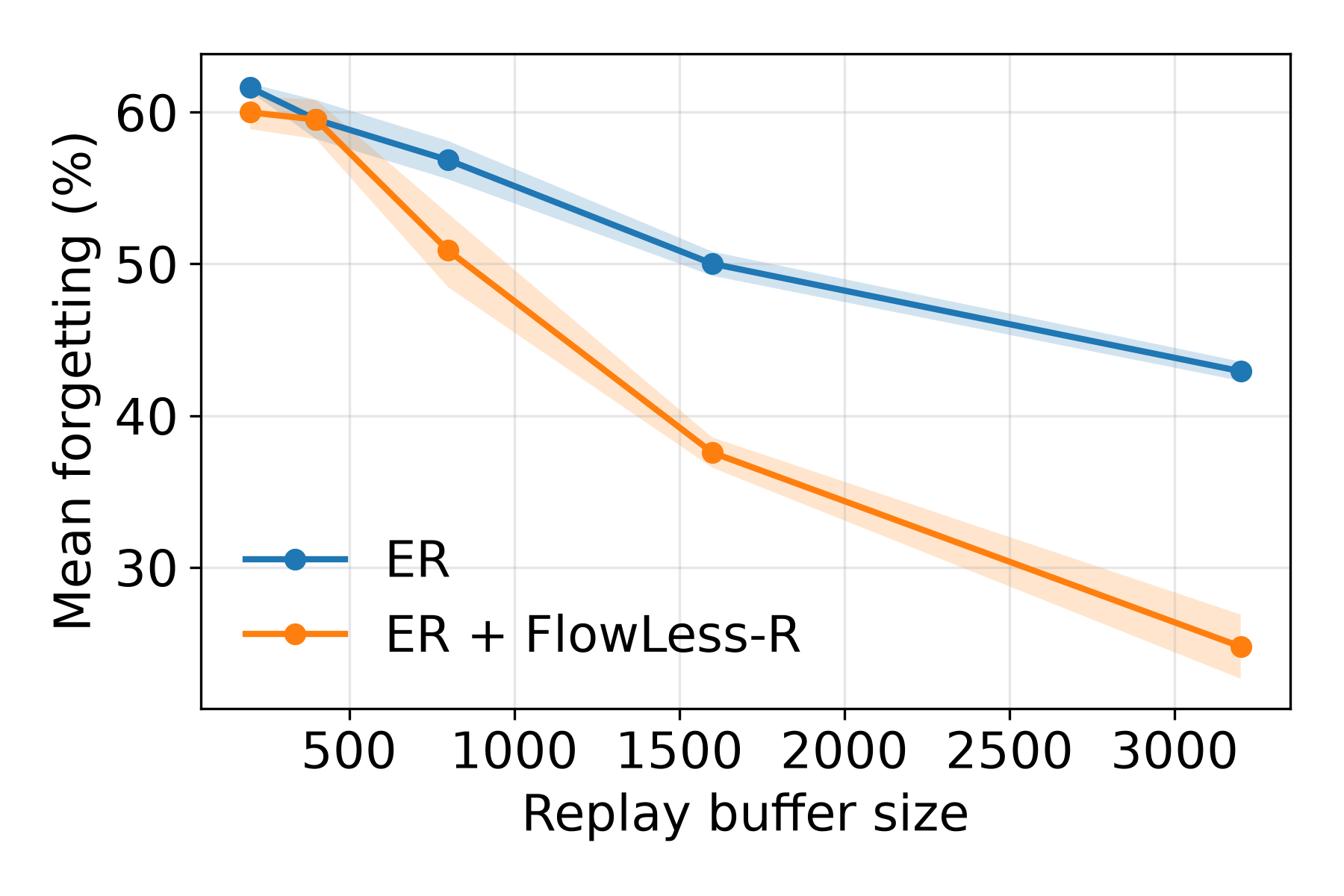}
        \caption{SplitTinyImageNet}
    \end{subfigure}

    \caption{Effect of replay buffer size on continual learning performance. The first row shows the final average accuracy, while the second row reports the mean forgetting. Curves represent the mean over random seeds, and shaded regions denote one standard deviation.}
    \label{fig:buffer_sweep}
\end{figure*}

\subsection{Effect of Density Weighting}
\label{sec:density_weighting}

FlowLess-R optionally weights replay samples according to their local representation density. Given the estimated density $\rho_i$ of replay sample $i$, the weight is defined as

\[
w_i=(\rho_i+\varepsilon)^{\alpha},
\]

where $\alpha$ controls the density-weighting strength and $\varepsilon$ is a small constant introduced for numerical stability. Negative values of $\alpha$ assign larger weights to samples located in low-density regions of the representation space, whereas positive values emphasize samples from high-density regions. The corresponding weighted regularizer is

\[
\mathcal{L}_{\mathrm{flux}}
=
\frac{\sum_i w_i \|z_i-\tilde z_i\|_2^2}
{\sum_i w_i}.
\]

To evaluate the sensitivity of FlowLess-R to the density-weighting exponent, we performed experiments on SplitMNIST using

\[
\alpha\in\{-3.33,-1,-0.333,-0.033,0,0.033,0.333,1,3.33\}.
\]

For each value of $\alpha$ and each replay memory size, the regularization coefficient $\lambda$ was selected to maximize the final average accuracy. The results are summarized in Table~\ref{tab:alpha}.
\begin{table}[!t]
\centering
\caption{Effect of the density-weighting exponent $\alpha$ on FlowLess-R
performance on SplitMNIST. For each value of $\alpha$ and replay memory size,
the regularization coefficient $\lambda$ was selected to maximize final average
accuracy. Results are reported as mean $\pm$ standard deviation over 10 matched
random seeds (0--9). For each memory size, differences in final average accuracy
across values of $\alpha$ were assessed using a Friedman test, with random seed
treated as the repeated-measures block. Pairwise comparisons between each
density-weighted configuration ($\alpha\neq0$) and the unweighted reference
($\alpha=0$) were performed using two-sided paired $t$-tests across matched
seeds, with Holm correction for multiple comparisons. Reported pairwise
$p$-values are Holm-adjusted.}
\label{tab:alpha}

\scriptsize
\setlength{\tabcolsep}{2pt}
\renewcommand{\arraystretch}{1.15}

\resizebox{\textwidth}{!}{
\begin{tabular}{c|cccc|cccc}
\hline
&
\multicolumn{4}{c|}{20 samples/task}
&
\multicolumn{4}{c}{80 samples/task}\\
\cline{2-9}

\rule{0pt}{4ex}$\alpha$
& $\lambda$
& Accuracy
& Forgetting
& $p$
& $\lambda$
& Accuracy
& Forgetting
& $p$\\
\hline

$-3.33$
& 3.0
& 65.28 $\pm$ 2.41
& 42.33 $\pm$ 3.03
& \textbf{$<10^{-4}$}
& 3.0
& 82.98 $\pm$ 1.39
& 20.01 $\pm$ 1.72
& \textbf{$<10^{-4}$}
\\

$-1.00$
& 1.0
& 70.83 $\pm$ 1.95
& 35.33 $\pm$ 2.44
& 1.000
& 3.0
& 86.74 $\pm$ 0.89
& 15.22 $\pm$ 1.07
& 1.000
\\

$-0.333$
& 1.0
& 70.70 $\pm$ 2.14
& 35.54 $\pm$ 2.71
& 1.000
& 1.0
& 86.70 $\pm$ 0.71
& 15.22 $\pm$ 0.92
& 0.792
\\

$-0.033$
& 0.3
& 70.35 $\pm$ 2.19
& 35.95 $\pm$ 2.75
& 0.123
& 1.0
& 86.59 $\pm$ 0.87
& 15.36 $\pm$ 1.12
& 1.000
\\

$\mathbf{0.000}$
& 0.3
& \textbf{70.97 $\pm$ 2.40}
& \textbf{35.19 $\pm$ 3.03}
& ---
& 1.0
& 86.54 $\pm$ 0.90
& 15.45 $\pm$ 1.13
& ---
\\

$0.033$
& 0.3
& 70.89 $\pm$ 2.29
& 35.31 $\pm$ 2.87
& 1.000
& 1.0
& \textbf{86.77 $\pm$ 0.94}
& \textbf{15.15 $\pm$ 1.18}
& 0.881
\\

$0.333$
& 1.0
& 70.34 $\pm$ 2.18
& 35.99 $\pm$ 2.74
& 0.467
& 1.0
& 86.47 $\pm$ 0.85
& 15.53 $\pm$ 1.09
& 1.000
\\

$1.00$
& 0.3
& 69.86 $\pm$ 2.69
& 36.61 $\pm$ 3.40
& 0.440
& 3.0
& 86.20 $\pm$ 1.02
& 15.98 $\pm$ 1.32
& 1.000
\\

$3.33$
& 3.0
& 66.51 $\pm$ 2.87
& 40.78 $\pm$ 3.59
& \textbf{0.0018}
& 3.0
& 84.88 $\pm$ 1.02
& 17.64 $\pm$ 1.27
& \textbf{$<10^{-4}$}
\\

\hline

\multicolumn{1}{c|}{\rule[-1.2ex]{0pt}{4ex}Friedman}
&
\multicolumn{4}{c|}{$p=7\times10^{-6}$}
&
\multicolumn{4}{c}{$p<10^{-6}$}
\\

\hline
\end{tabular}
}
\end{table}
FlowLess-R is largely insensitive to the choice of the density-weighting
exponent over a broad range of values. Across replay memories of 20 and 80
samples per task, the differences in final average accuracy and forgetting
remain small for $-1\leq\alpha\leq1$. Although the Friedman test for repeated
measures detects statistically significant differences among the evaluated
values of $\alpha$ ($p=7\times10^{-6}$ for 20 samples per task and
$p<10^{-6}$ for 80 samples per task), Holm-adjusted paired $t$-tests show
that significant pairwise differences from the unweighted formulation
($\alpha=0$) are observed only for the extreme values $\alpha=\pm3.33$,
which yield significantly lower performance. No statistically significant
differences are observed between $\alpha=0$ and any value in the range
$-1\leq\alpha\leq1$. These results suggest that density weighting does not
provide a measurable benefit over the unweighted formulation, whereas
excessively large positive or negative weighting exponents degrade
performance.

Consequently, in the rest of this work, we use the unweighted formulation ($\alpha=0$), which performs comparably to moderate density-weighting schemes while avoiding an additional hyperparameter.

\clearpage
\section{Miscellaneous}

\paragraph{Reproducibility.}
Source code for reproducing the experiments, analyses, and figures presented
in this work is publicly available at
\url{https://github.com/maksimkazanskii/FlowLess}.

\paragraph{Large language model usage.}
ChatGPT (OpenAI; GPT-5.6 Sol) was used to assist with language editing and software development. All LLM-assisted outputs were reviewed and verified by the author, who takes full responsibility for the content of the paper.

\end{document}